\documentclass[acmsmall]{acmart}
\AtBeginDocument{%
  }

\setcopyright{cc}
\setcctype{by}
\acmDOI{10.1145/3808143}
\acmYear{2026}
\acmJournal{PACMSE}
\acmVolume{3}
\acmNumber{FSE}
\acmArticle{FSE136}
\acmMonth{7}
\acmSubmissionID{fse26mainb-p519-p}
\received{2026-01-23}
\received[accepted]{2026-03-24}

\usepackage{microtype}
\usepackage{graphicx}
\usepackage{subcaption}
\usepackage{booktabs} 
\usepackage{tabularx}
\usepackage{hyperref}
\usepackage{mdframed}
\usepackage[dvipsnames, svgnames]{xcolor}
\usepackage{caption}  
\usepackage{wrapfig}
\usepackage[linesnumbered,ruled,vlined]{algorithm2e}
\usepackage{ulem}

\usepackage[most]{tcolorbox}
\tcbuselibrary{skins,xparse,breakable}

\newcommand{\method}{StepFly}

\begin{document}

\title{StepFly: Agentic Troubleshooting Guide Automation for Incident Diagnosis}

\author{Jiayi Mao}
\orcid{0009-0000-5488-7813}
\affiliation{%
  \institution{Tsinghua University}
  \city{Beijing}
  \country{China}
}
\email{maojy23@mails.tsinghua.edu.cn}

\author{Liqun Li}
\orcid{0000-0003-4579-3799}
\affiliation{%
  \institution{Microsoft}
  \city{Beijing}
  \country{China}
}
\email{liqun.li@microsoft.com}

\author{Yanjie Gao}
\orcid{0000-0003-1899-8561}
\affiliation{%
  \institution{Microsoft Research}
  \city{Beijing}
  \country{China}
}
\affiliation{%
  \institution{Renmin University of China}
  \city{Beijing}
  \country{China}
}
\email{yanjga@microsoft.com}

\author{Zegang Peng}
\orcid{0009-0003-0325-1593}
\affiliation{%
  \institution{Tsinghua University}
  \city{Beijing}
  \country{China}
}
\email{pzg24@mails.tsinghua.edu.cn}

\author{Shilin He}
\orcid{0000-0002-8595-5388}
\affiliation{%
  \institution{Microsoft}
  \city{Beijing}
  \country{China}
}
\email{selin.hsl@gmail.com}

\author{Chaoyun Zhang}
\orcid{0000-0002-1304-6839}
\affiliation{%
  \institution{Microsoft}
  \city{Beijing}
  \country{China}
}
\email{chaoyun.zhang@microsoft.com}

\author{Si Qin}
\orcid{0000-0002-8698-1860}
\affiliation{%
  \institution{Microsoft}
  \city{Beijing}
  \country{China}
}
\email{Si.Qin@microsoft.com}

\author{Samia Khalid}
\orcid{0009-0009-9893-594X}
\affiliation{%
  \institution{Microsoft}
  \city{Redmond}
  \country{USA}
}
\email{sakhal@microsoft.com}

\author{Qingwei Lin}
\orcid{0000-0003-2559-2383}
\affiliation{%
  \institution{Microsoft}
  \city{Beijing}
  \country{China}
}
\email{qlin@microsoft.com}

\author{Saravan Rajmohan}
\orcid{0000-0002-2019-213X}
\affiliation{%
  \institution{Microsoft}
  \city{Redmond}
  \country{USA}
}
\email{saravan.rajmohan@microsoft.com}

\author{Sitaram Lanka}
\orcid{0009-0008-7545-9461}
\affiliation{%
  \institution{Microsoft}
  \city{Redmond}
  \country{USA}
}
\email{slanka@microsoft.com}

\author{Dongmei Zhang}
\orcid{0000-0002-9230-2799}
\affiliation{%
  \institution{Microsoft}
  \city{Beijing}
  \country{China}
}
\email{dongmeiz@microsoft.com}

\renewcommand{\shortauthors}{Mao and Li, et al.}

\begin{abstract}
Effective incident management in large-scale IT systems relies on troubleshooting guides (TSGs), but their manual execution is slow and error-prone. 
While recent advances in LLMs offer promise for automating incident management tasks, existing LLM-based solutions lack specialized support for several key challenges, including managing TSG quality issues, interpreting complex control flow, handling data-intensive queries, and exploiting execution parallelism. 
We first conducted an empirical study on 92 real-world TSGs, and, guided by our findings, we present \method, a novel end-to-end agentic framework for troubleshooting guide automation. 
Our approach features a three-stage workflow: the first stage provides a comprehensive guide together with a tool, TSG Mentor, to assist site reliability engineers (SREs) in improving TSG quality; the second stage performs offline preprocessing using LLMs to extract structured execution directed acyclic graphs (DAGs) from unstructured TSGs and to create dedicated Query Preparation Plugins (QPPs); and the third stage executes online using a DAG-guided scheduler-executor framework with a memory system to ensure correct workflow and support parallel execution of independent steps.
Our empirical evaluation on a collection of real-world TSGs and incidents demonstrates that \method~ achieves a $\sim$94\% success rate on GPT-4.1, outperforming baselines with less time and token consumption. Furthermore, it achieves a remarkable execution time reduction of 32.9\% to 70.4\% for parallelizable TSGs. Our code and sample data are publicly available at \url{https://github.com/microsoft/StepFly}.
\end{abstract}



\maketitle

\section{Introduction}\label{sec:intro}

Incident management (IcM) is a core process in large-scale IT service operations that aims to minimize service disruptions and meet strict service-level objectives (SLOs). Troubleshooting guides (TSGs) underpin this process by providing step-by-step instructions that site reliability engineers (SREs) use to triage, diagnose, mitigate, and resolve incidents \cite{chen_icm_2020}. A TSG is an unstructured document that captures recommended procedures, decision points, and best practices for a specific class of incidents. 
The effective use of a TSG requires an SRE to rapidly internalize the guide, adapt its generic instructions to the live incident context, and execute each step without omitting critical details. Manual execution under time pressure can be slow and error-prone, leading to misinterpreted steps, missed prerequisites, and delayed decision branching, which ultimately increases the time to resolution.

Recent advances in large language models (LLMs) show promise for automating incident-management tasks, including triage \cite{zexin2024comet}, query generation \cite{yuxuan2024xpert}, and decision support \cite{pengxiang2023summary, chen2024automatic}. Prior work has also applied LLM-based agents to TSG execution \cite{an_nissist_2024, las-casas_llexus_2024, zhang2024flash}. However, these efforts largely treat TSGs as well-structured documentation with consistently high-quality, and rely on generic agent designs, lacking specialized considerations for the core challenges in achieving reliable and efficient TSG automation.

In this work, we first conducted a comprehensive study of TSG documentation to understand its characteristics and challenges for automation. TSGs are often documentation for lengthy, complex tasks that require deep domain knowledge.
Unlike code in programming languages \cite{stroustrup1986overview, eckel2003thinking, matsakis2014rust}, TSGs are unstructured and lack type constraints \cite{cardelli1996type, pierce2002types}, which makes them difficult for LLM-based agents to comprehend. Furthermore, SREs often create TSGs as sparse notes, omitting crucial implicit knowledge and lacking clarity. As a result, we first developed a TSG writing guidance to help SREs to improve clarity. We also introduce TSG Mentor, a novel tool that automatically analyzes and reformulates TSGs while offering feedback to improve their quality. Our empirical evaluation showed the tool's effectiveness in detecting common issues, achieving a recall of 0.78, a precision of 0.85, and an F1-score of 0.81.

Despite TSG quality issues, achieving robust and efficient TSG execution is a major challenge due to several factors. First, TSGs have complex control flow, including conditional branching and early termination, which is difficult for LLMs to handle. Furthermore, they often involve large, templated queries that are prone to generation and execution errors, and require interaction with various data sources for data-intensive analytics. Finally, the sequential execution of TSGs, which are time-critical, introduces delays and increases the risk of errors, particularly in complex scenarios. 

To address these challenges, we propose \textbf{\method}, a framework that involves offline preprocessing and online execution.
In the offline stage, we use LLMs to automatically extract structured execution DAGs from raw TSGs. These DAGs precisely capture control flow and step dependencies, guiding the entire execution process. We also extract Query Preparation Plugins (QPPs), which encapsulate query templates and parameters. This design eliminates the need for in-context query generation, making queries more consistent and reducing errors. We leverage LLMs for all preprocessing, and our evaluation shows high precision results and requires minimal human intervention.

Our online execution stage features two key designs: a DAG-guided execution engine and a memory system. The execution engine uses a scheduler-executor architecture to manage step execution based on dependencies defined in the DAG. This ensures the agent adheres strictly to the workflow, minimizing planning errors like skipping steps or executing them out of order. Furthermore, the scheduler can allocate multiple executors to independent steps, enabling parallel execution and improving overall efficiency. The memory system allows the agent to retain and manage structured data from plugins, preventing it from being treated as unstructured text. This design significantly improves accuracy and efficiency. Our code and synthesized incident data are publicly available at \url{https://github.com/microsoft/StepFly}.

We conducted experiments on real-world TSGs and incidents to evaluate our proposed framework. The results show significant improvements in both TSG execution efficiency and accuracy.
On a powerful LLM like GPT-4.1, \method \ achieved a success rate of $\sim$94\%, which significantly outperforms baseline methods. Even with a more modest model like GPT-4.1-mini, our framework maintained a high success rate of $\sim$84\%.
Our evaluations on token and time consumption indicate that \method~ costs much less compared to other methods. Additionally, on a subset of TSGs that can be parallelized, we demonstrated a further reduction of execution time by 32.9\% to 70.4\%, confirming the value of our parallel execution design.
Our main contributions are as follows: 
\begin{itemize}
\item We conducted a comprehensive empirical study of real-world TSGs to identify common challenges in documentation and execution. Based on our findings, we introduce TSG Mentor, a novel tool that automatically provides feedback to improve TSG quality.
\item We introduce \method \ -- a novel LLM-powered agent framework for automated TSG execution. We evaluated our framework on real-world incidents, achieving a success rate of $\sim$94\% on GPT-4.1 and $\sim$84\% on GPT-4.1-mini, which significantly outperforms baseline methods, yet with less execution time and token consumption.
\item We demonstrate that \method \ naturally supports the parallelization of independent TSG steps. Our approach achieves a speedup that reduces execution time by 32.9\% to 70.4\% for a set of parallelizable TSGs.
\end{itemize}

\section{Background}\label{sec:background}
In this section, we introduce background on incident management and TSGs.

\subsection{Incident Management}\label{subsec:icm}

Incident management (IcM) refers to the process of handling live-site issues reported by either monitoring systems or customers. 
Timely incident mitigation is critical for ensuring the availability of cloud systems or online services, thereby reducing customer impact \cite{wang2021how}. 
Today, the vast majority of incidents are reported by automated monitoring systems in large companies~\cite{chen_icm_2020}. 
When an incident is detected, a common practice for handling incidents is for SREs to follow a predefined TSG, who are responsible for resolving the incident. 

\subsection{Troubleshooting Guide}\label{subsec:tsg}

TSGs are essential tools in incident management, providing a step-by-step approach to issue resolution. While the format and style presentation of TSGs can vary, their underlying structure is generally consistent.

Typically, each step within a TSG includes: (1) a concise title summarizing the action; (2) a detailed description, which may include instructions, commands, queries, etc.; (3) a specification of expected outcomes; and (4) flow control logic that dictates the subsequent step based on the observed outcomes.
A TSG terminates at one or more \textit{termination points}. It's important to note that reaching a termination point does not necessarily indicate issue resolution; it may also indicate that the problem lies outside the TSG's defined scope.

\begin{wrapfigure}{l}{0.45\textwidth}
    \centering
    \includegraphics[width=\linewidth]{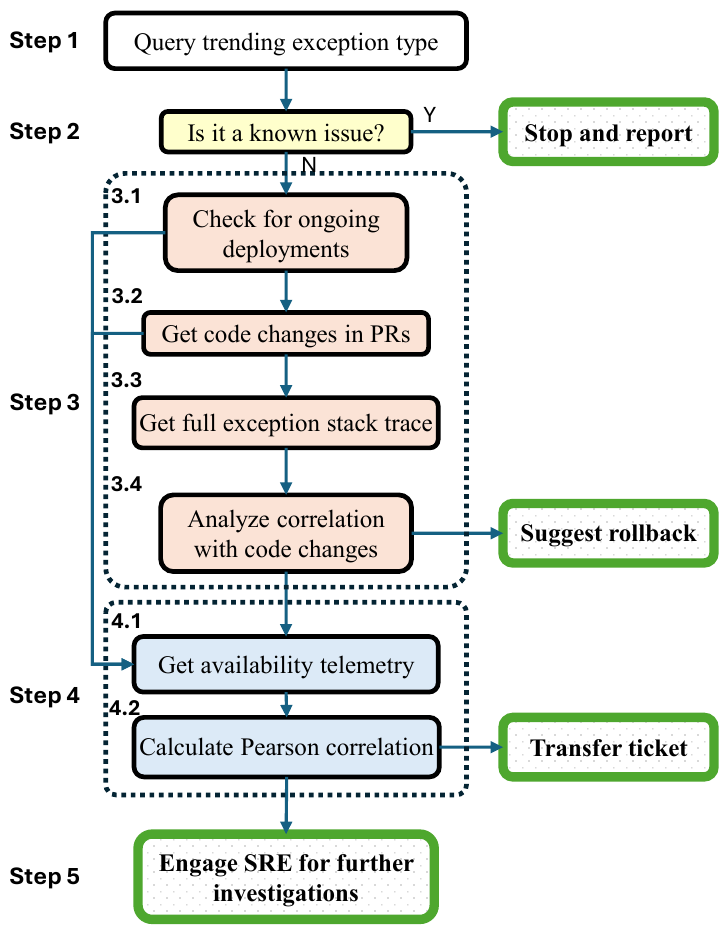}
    \caption{Workflow of a real TSG for diagnosing availability incidents of a large online service.}
    \label{fig:example_tsg}
    \vspace{-5pt}
\end{wrapfigure}

Figure \ref{fig:example_tsg} illustrates a real TSG workflow for diagnosing a class of availability drop incidents. The process begins with Step 1, which queries for the top trending exception type from service logs. This result is then checked in Step 2 against a list of known exceptions. If it’s a known issue, this reaches a termination point, and the SRE can stop. If not, they proceed to Step 3 to check if the drop was caused by code changes. 
Step 3 consists of four sub-steps: checking for ongoing deployments (3.1), getting relevant code changes (3.2), retrieving the full exception stack from logs (3.3), and analyzing the correlation of the exception stack with the code changes (3.4).
If the exception is raised in a recently changed file, the recommended action is a rollback. Otherwise, the SRE moves to Step 4. The process also branches directly to Step 4 if no ongoing deployments or relevant code changes are found in sub-steps 3.1 and 3.2.
Step 4 checks if the issue is caused by an upstream dependent service. This involves two sub-steps: getting availability metrics for both services (4.1) and calculating the Pearson correlation (4.2). If the correlation is high, the incident ticket is transferred to the upstream team. If not, the diagnosis ends without a decisive conclusion, and an SRE is engaged for further investigation. The four termination points are highlighted with green, thick-bordered rectangles.

TSGs, authored by experienced SREs, are the primary guides for resolving incidents. When an incident occurs, an SRE meticulously follows the steps in the relevant TSG. This process requires the SRE to interact with various data sources and perform analysis. For example, Step 1 of the TSG in Fig. \ref{fig:example_tsg} directs the SRE to run a complex query (51 lines) to find the trending exception type during the time of the incident. This involves several manual actions: extracting parameters (11 required parameters in this example) from the incident report, filling them into the query template, and running the query on a database explorer. After getting the results, the SRE must interpret them, often by writing code or using tools, to decide on the next action. 

Even with ready TSGs, manual execution is a significant bottleneck, consuming valuable time and potentially delaying issue resolution, especially for large online services. This inherent inefficiency strongly motivates the need for automation. A recent study \cite{yuxuan2024xpert} highlights that preparing queries is among the most frequent and time-consuming tasks for SREs, further underscoring the benefits of automating these procedures within TSGs. 
Furthermore, we've found that TSG quality is itself a challenge that hinders efficiency in both manual and automated execution, a topic we will discuss in the next section.

\section{Empirical Study on TSGs}\label{subsec:tsg_scope}

In this study, we focus on a subset of 92 TSGs from 9 different teams in the company. We selected this scope based on three criteria: (1) the TSGs have been applied to a high frequency of incidents in recent months; (2) the involved teams manage high-traffic services, providing strong motivation for TSG automation; and (3) the selection covers a broad range of diagnostic complexities, spanning from simple 3-step procedures to complex 30-step workflows. This selection ensures our dataset is robust and representative.

We restricted our study to TSGs that rely solely on common data sources such as service logs, metrics, and DevOps platforms. Our rationale was to focus on procedures that are more readily automatable, given the complexity of authenticating and managing specialized plugins. 

\subsection{Characteristics and Structure}\label{subsec:tsg_characteristics}

\begin{figure}[t]
    \centering
    \begin{subfigure}{0.244\columnwidth}
        \centering
        \includegraphics[width=\linewidth]{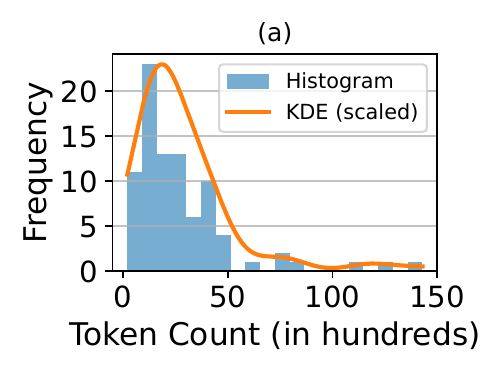}
        \label{fig:token_dist}
    \end{subfigure}
    \hfill 
    \begin{subfigure}{0.244\columnwidth}
        \centering
        \includegraphics[width=\linewidth]{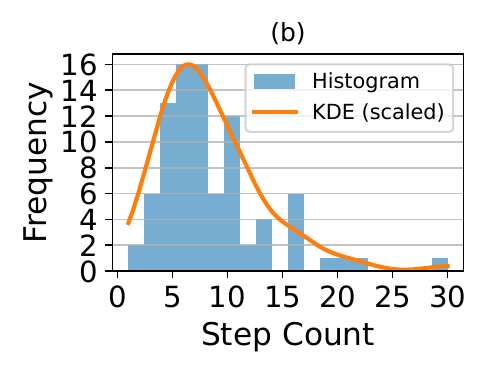}
        \label{fig:step_dist}
    \end{subfigure}
    \hfill
    \begin{subfigure}{0.244\columnwidth}
        \centering
        \includegraphics[width=\linewidth]{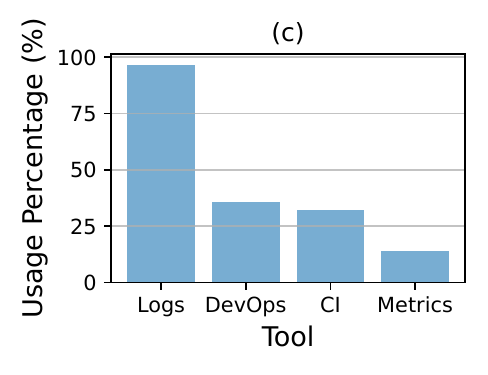}
        \label{fig:plugin_usage}
    \end{subfigure}
    \hfill
    \begin{subfigure}{0.244\columnwidth}
        \centering
        \includegraphics[width=\linewidth]{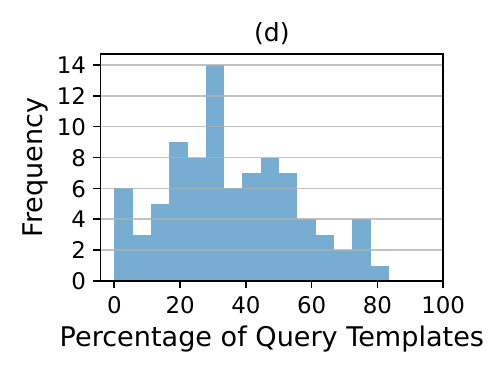}
        \label{fig:kusto_dist}
    \end{subfigure}
    \vspace{-10pt}
    \caption{Statistics on TSG characteristics: (a) Token count distribution, (b) Step count Distribution, (c) Tool usage, (d) Query template percentage.}
    \label{fig:combined_distributions}
\end{figure}

Our analysis of the TSG characteristics, as shown in Fig. \ref{fig:combined_distributions} (a) and (b), reveals key insights into their size and structure. The token distribution (a) shows that most TSGs contain approximately 3K tokens (with the GPT-4o tokenizer), while some outliers can exceed 10K tokens. In terms of structure, our analysis reveals that the majority of TSGs in our dataset consist of 5-15 steps, with some complex guides reaching up to 30 steps. For example, the TSG illustrated in Fig. \ref{fig:example_tsg} contains 9 steps or sub-steps.

\begin{tcolorbox}[colframe=gray!10!white, colback=gray!10!white,
    shadow={1mm}{-1mm}{2mm}{gray!40!white}]
    \textbf{Finding 1:} TSGs are often lengthy documents with many steps and complex conditional connections. This inherent complexity makes it challenging for LLMs to follow the correct execution path \cite{li2025longcontext, liu2025contextgaps, NEURIPS2024_babilong}, as confirmed by our experiments in Sec.~\ref{subsec:rq2}, where baselines often fail to navigate through the steps correctly.
\end{tcolorbox}

Upon reviewing the TSGs, a noteworthy characteristic is the independence of many of their steps.
In the example TSG from Fig. \ref{fig:example_tsg}, the SRE must assess different possibilities: a known issue, code changes, or an upstream service problem. All these assessments could be performed concurrently, especially when a team of SREs is collaborating. Although authors often present these steps sequentially, this may not be a strict requirement. 

\begin{wrapfigure}{r}{0.25\textwidth}
    \centering
    \includegraphics[width=\linewidth]{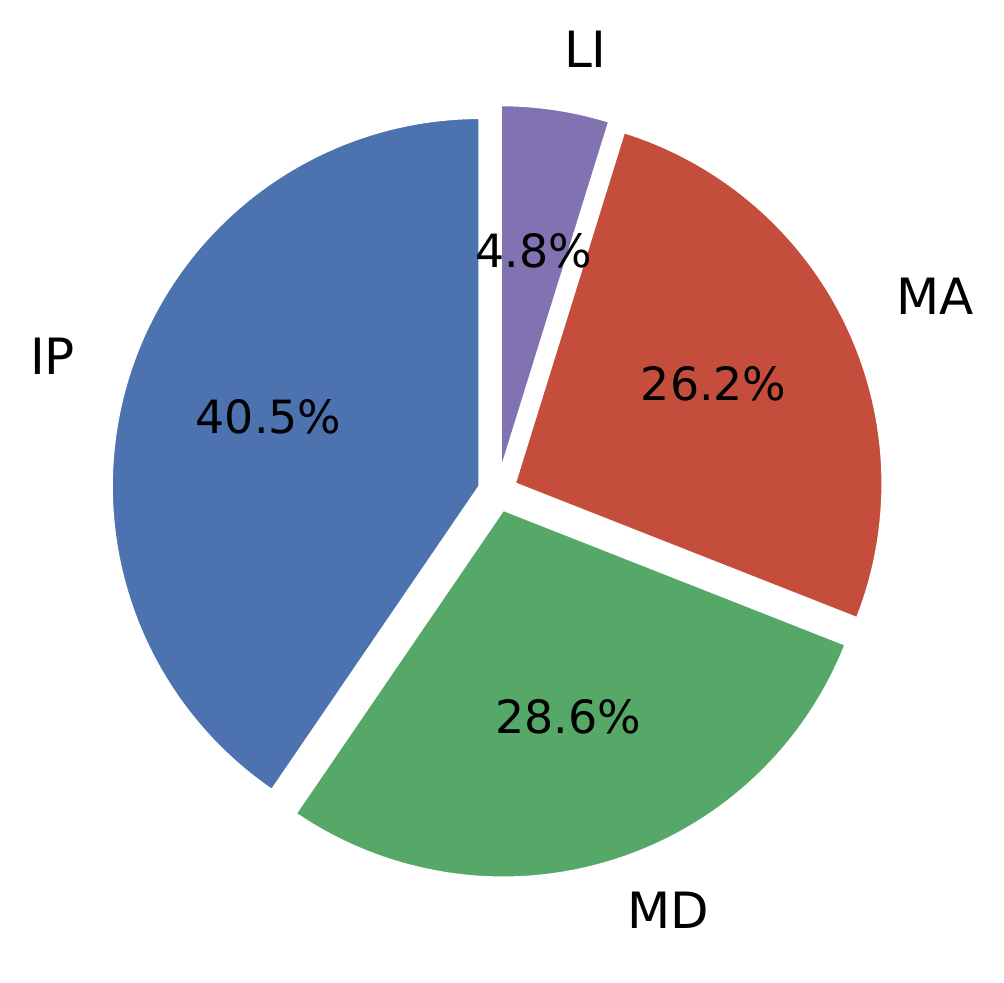}
    \caption{TSG parallelization opportunity categories.}
    \label{fig:parallelization_opportunities}
    \vspace{-5pt}
\end{wrapfigure}

We systematically analyzed the TSG corpus for parallelization opportunities, defined by an absence of data dependencies and sequential constraints between steps. Our analysis reveals that $\sim$46\% of the TSGs in our study exhibit potential for concurrency, distributed across four categories, shown in Fig. \ref{fig:parallelization_opportunities}: (1) \textit{Independent Paths} (IP, $\sim$40.5\%), representing diagnostic branches—such as known issues and service health checks—that can be explored concurrently; (2) \textit{Multiple Data Sources} (MD, $\sim$28.6\%), involving parallel queries to disparate data sources; (3) \textit{Multiple Analysis Types} (MA, $\sim$26.2\%), comprising different analytical dimensions of the same data processed simultaneously; and (4) \textit{Loop Iterations} (LI, $\sim$4.8\%), involving repeated queries with varying parameters. The prevalence of these categories across nearly half of the analyzed TSGs confirms that parallelization is a pervasive characteristic. 

This observation offers a significant opportunity to reduce diagnosis latency through the parallel execution of independent steps, which we shall discuss further in Sec. \ref{subsec:parallelization}.

\begin{tcolorbox}[colframe=gray!10!white, colback=gray!10!white, 
    shadow={1mm}{-1mm}{2mm}{gray!40!white}]
    \textbf{Finding 2:} The structural independence of many TSG steps offers a significant opportunity to enhance the efficiency of automated execution through parallel processing, inspired by how a team of SREs would collaborate.
\end{tcolorbox}

\subsection{Tool Usage and Query Templates}\label{subsec:tool_usage}

This section aims to understand the tools involved in incident diagnosis by following TSGs, corresponding to the tools that will be presented for the agent.

As shown in Fig. \ref{fig:combined_distributions} (c), service log data retrieval is the most frequent operation for SREs. 
At Microsoft, service logs are stored in a high-performance database based on the Kusto engine \cite{kusto_kql}, organized in tables with numerous columns. 
SREs can use the Kusto Query Language (KQL) to write complex queries that often include table joins, filters, and aggregations.
These queries are typically provided as templated code blocks in the TSGs for manual execution. An SRE can copy the template into a data explorer, populate it with parameters from the incident report, and execute the query. 

As shown in Fig. \ref{fig:combined_distributions} (d), these query templates constitute, on average, $\sim$35.85\% of the total tokens per TSG. We identified a total of 405 query templates across the TSGs in our study, averaging 4.4 queries per TSG. These queries exhibit significant complexity, with a mean length of 26.6 lines of code and a maximum of 109 lines. This reflects the widespread adoption of this log storage engine within Microsoft, also observed in a recent work \cite{yuxuan2024xpert}.
Besides service logs, metric data, such as service availability or latency, is another commonly used data source for both incident detection and diagnosis, as seen in Step 4 of the example TSG in Fig. \ref{fig:example_tsg}.

SREs use DevOps tools to check for ongoing deployments and find relevant pull requests (PRs) or code changes, as illustrated in Step 3 of the example TSG in Fig. \ref{fig:example_tsg}.
While other tools are explicitly referenced in TSGs, the Code Interpreter (CI) is often an unstated but necessary tool. After obtaining various incident data, an SRE may write code or use tools like spreadsheets to perform simple data analysis. We consider all such operations that can be handled with a code interpreter. Examples include estimating correlations between time-series data (Step 4.2 in Fig. \ref{fig:example_tsg}), calculating data distributions, and detecting significant metric fluctuations.

Unlike tools that accept simple parameters (e.g., numeric or categorical values), service log retrieval requires the synthesis of complete queries. When automating the TSG, the LLM must generate these queries dynamically based on the context (query templates and incident information). Our experiments identify this ``on-the-fly'' generation as a primary source of error, characterized by: (1) \textit{instruction drift}, where the LLM ignores templates to rewrite queries; (2) \textit{structural omissions}, such as missing sub-queries or conditions; and (3) \textit{syntax errors}, particularly incorrect regex formatting due to escape character mishandling. Similar challenges in LLM-driven KQL generation are documented in recent literature \cite{hallucinationkql25,tang2024nl2kql}.

\begin{tcolorbox}[colframe=gray!10!white, colback=gray!10!white, 
    shadow={1mm}{-1mm}{2mm}{gray!40!white}]
    \textbf{Finding 3:} The dominant presence of query templates poses two primary challenges for LLM-based automation: incorrect query generation for complex or lengthy queries and inefficiency due to the considerable time and tokens required for their generation.
\end{tcolorbox}

Our evaluation in Sec.~\ref{subsec:rq2} identifies query generation issues as a primary bottleneck.
Beyond reliability, dynamic generation imposes substantial overhead; query verbosity consumes significant output tokens and exacerbates latency, as detailed in our case study in Sec.~\ref{subsec:case2}. Collectively, these reliability and efficiency concerns constitute the two primary hurdles for automated TSG execution.

\subsection{TSG Quality Issues}\label{subsec:tsg_quality}

In this work, we focus on TSG quality from the perspective of an LLM agent for automation, though we note that most of the issues are equally applicable to human readers. 
We conducted a manual labeling study across our selected TSGs. Our labeling team consisted of four annotators (two researchers and two SREs), who first calibrated on 10 pilot TSGs using a detailed annotation guideline, examples, and decision criteria. Each of the 92 TSGs was then independently reviewed by two annotators performing line-by-line labeling, with researcher-SRE pairing when possible. We measured inter-rater reliability using Cohen's kappa, achieving $\kappa = 0.78$ (substantial agreement)~\cite{landis1977measurement}, with the highest agreement on DI issues ($\kappa = 0.85$, almost perfect) due to their concrete nature, and the lowest on CP issues ($\kappa = 0.72$, substantial) due to inherent subjectivity. Disagreements were resolved through discussion, with $\sim$12\% requiring adjudication by a third annotator.
The identified issues are empirically categorized into five groups, as presented in Table \ref{tab:tsg_issues}.

\begin{table}[h]
\vspace{-5pt}
\centering
\caption{Categories of TSG Issues.}
\label{tab:tsg_issues}
\scalebox{0.79}{
\begin{tabular}{p{4.4cm} p{11cm}}
\toprule
\textbf{Category} & \textbf{Description} \\
\midrule
Clarity and Precision (CP) & Issues related to the clarity and precision of instructions, including ambiguous references or non-actionable steps. \\
Control Flow (CF) & Issues concerning the logical flow of the TSG, such as wrong or unclear next steps. \\
Data Flow (DF) & Issues involving the handling of data within the TSG, such as missing parameters or no description of data schema. \\
Database Instruction (DI) & Issues specific to database queries, particularly log queries, including syntax errors or missing connection information. \\
Presentation and Structure (PS) & Issues related to the overall presentation and structure of the TSG, such as formatting problems or unmarked termination points. \\
\bottomrule
\end{tabular}
}
\vspace{-5pt}
\end{table}

\begin{wrapfigure}{r}{0.25\textwidth}
    \centering
    \includegraphics[width=\linewidth]{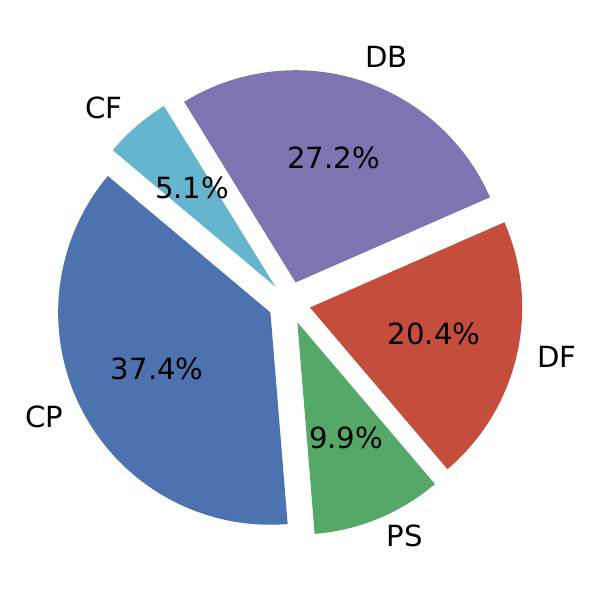}
    \caption{TSG Issue Distribution.}
    \label{fig:issue_dist}
\end{wrapfigure}

Each major issue category is further decomposed into sub-categories. For example, the Data Flow Issues (DF) category includes sub-issues such as ``Unknown Input Source'', ``Wrong Input Source'', and ``Missing Parameters''; the Control Flow Issues (CF) group includes sub-issues such as ``Unable To Infer Next Step'' and ``Wrong Next Step''. The detailed issue taxonomy is omitted due to space constraints.
Figure \ref{fig:issue_dist} shows the distribution of issues.


Clarity and Precision (CP) issues are the most common in our study, accounting for 37.4\% of all issues; the dominant sub-issues are ``Missing Description of the Action'' and ``Unquantifiable Condition'', reflecting insufficient execution guidance and hard-to-assess conditions.
Database Instruction (DI) issues rank second at 27.2\%, most often due to query templates with hardcoded parameters such as time ranges.
Data Flow (DF) and Control Flow (CF) issues account for 20.4\% and 5.1\%, respectively, with ``Unknown Input Source'' and ``Unable To Infer Next Step'' as the most common sub-issues.
Presentation and Structure (PS) issues make up the remaining 9.9\%, primarily due to ``Formatting Issue'', indicating deficiencies in layout and organization.

\begin{tcolorbox}[colframe=gray!10!white, colback=gray!10!white, 
    shadow={1mm}{-1mm}{2mm}{gray!40!white}]
    \textbf{Finding 4:} Most TSGs, in their current form, are not readily suitable for automation due to various quality issues. These issues require substantial preprocessing or refinement before automation can be effectively implemented.
\end{tcolorbox}

\section{The Proposed Approach}\label{sec:approach}
\begin{figure}[th]
\centering
\includegraphics[width=.88\columnwidth]{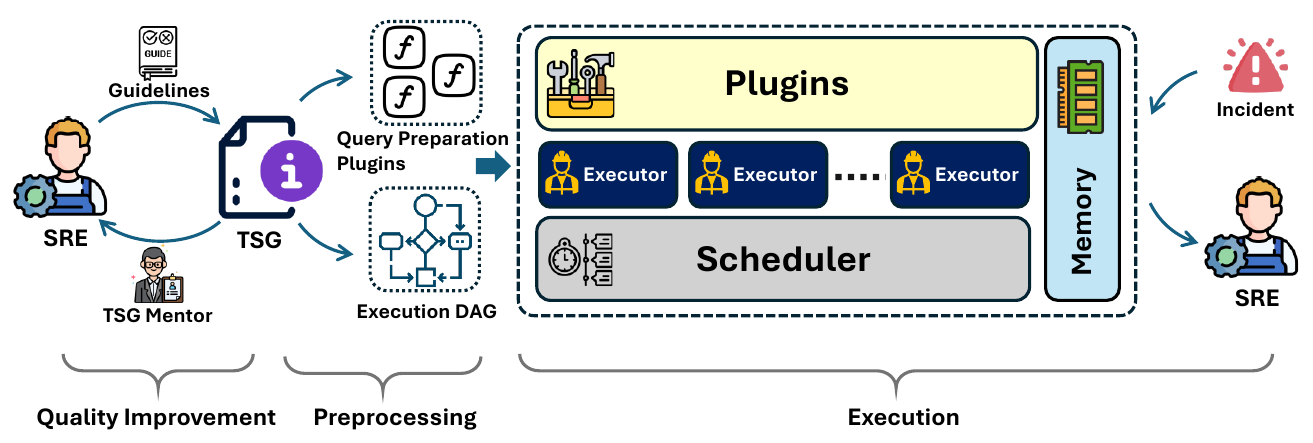}
\caption{The \method{} framework: a three-stage approach for TSG automation comprising quality improvement, preprocessing, and online execution.}
\label{fig:approach}
\end{figure}

In this section, we present our approach, \method{}, for TSG automation, which is motivated by the findings discussed in the previous section. \method{} is an end-to-end framework comprising three stages: (1) \textit{TSG quality improvement} (Sec.~\ref{subsec:guidelines}), which introduces quality guidelines and the TSG Mentor tool to refine TSG content; (2) \textit{Preprocessing} (Sec.~\ref{subsec:execution_flow}--\ref{subsec:plugin_extraction}), which extracts the execution DAGs and Query Preparation Plugins (QPPs) to bridge the gap between static instructions and executable logic; and (3) \textit{TSG execution} (Sec.~\ref{subsec:agent}--\ref{subsec:parallelization}), which employs a DAG-guided scheduler-executor architecture, to perform online diagnosis. The first two stages are performed offline, while the final stage is executed online. The overall architecture is shown in Fig. \ref{fig:approach}.

Motivated by Finding 4, we provide detailed guidance on writing high-quality TSGs, based on our empirical study in Section \ref{subsec:tsg_quality}. We introduce a tool called TSG Mentor to help SREs identify quality issues in their TSGs. The specifics are discussed in Section \ref{subsec:guidelines}.

We preprocess TSGs by creating an execution DAG, which serves as a structural blueprint of the workflow, with steps as nodes and their connections as edges. Since query templates constitute a significant component of TSGs, we extract them and create dedicated Query Preparation Plugins (QPPs). This approach eliminates the need for LLMs to generate queries on the fly. We discuss these preprocessing steps further in Sections \ref{subsec:execution_flow} and \ref{subsec:plugin_extraction}. 

The final stage is TSG execution, which is performed online by a novel scheduler-executor agent system. As shown in Fig. \ref{fig:approach}, the scheduler orchestrates the execution of TSG steps, guided by the DAG obtained from offline preprocessing. The executors, in turn, are responsible for carrying out these steps. A memory system is used to store and manage the data exchanged throughout the process, ensuring it remains structured rather than isolated text messages. Finally, plugins facilitate interactions with external systems, such as querying service logs and metrics, and performing DevOps operations. The full details of the TSG execution system are discussed in Section \ref{subsec:agent}.

With the proposed approach, upon the occurrence of an incident, the agent first loads the corresponding TSG (including the QPPs and DAG) and then follows the steps to diagnose the incident. When this process finishes, a final conclusion will be sent to SRE for review.

\subsection{TSG Quality Improvement}\label{subsec:guidelines}
In this section, we will discuss the TSG quality improvement guidelines and the tool we developed to automatically detect the TSG issues.

\paragraph{TSG Quality Improvement Guidelines.}
We developed comprehensive TSG quality guidelines to help authors create high-quality, actionable documentation based on our empirical study. These guidelines address various aspects of TSG writing, such as clarity, control and data flow, and structure.
A key component is a template for each step, which includes a title, instructions, and explicit connections to all possible next steps and their conditions. 
The guidelines also provide a checklist of the most common issues that we identified in Sec. \ref{subsec:tsg_quality}, offering detailed descriptions and examples. Our goal is to ensure TSGs are well-structured and easy to understand, thereby improving overall documentation quality.
Notably, these guidelines benefit not only LLM-based agents but also human readers.

\paragraph{TSG Mentor Tool.}
We developed \textit{TSG Mentor} to help authors follow the quality guidelines. The tool uses an LLM with a specialized prompt that includes the quality guidelines, examples of known issues, and sample TSGs. It classifies issues into the predefined categories listed in Table~\ref{tab:tsg_issues}, automatically formats the document, and provides inline annotations with suggestions for improvement. For evaluation, we used a leave-one-out strategy on our entire dataset of 92 manually labeled TSGs from Sec.~\ref{subsec:tsg_scope}, which collectively contain quality issues identified in our empirical study.
The input was the original TSG, and the output was a reformatted TSG with inline annotations of identified issues. We considered a successful match to be the same issue category and within $\pm$5 lines.
Due to the LLM's limited context, we dynamically selected the three most similar TSGs as few-shot examples based on their vector distances. This setup achieved a recall of 0.78, a precision of 0.85, and an F1 score of 0.81, demonstrating that TSG Mentor can effectively identify issues and assist in quality improvement.

A critical requirement in our revision process was maintaining human readability alongside LLM compatibility. We explicitly required that TSGs remain accessible and comprehensible to human SREs, not exclusively optimized for automated agents. This dual-compatibility approach reflects the practical reality that SREs will continue to use these TSGs for manual troubleshooting and maintenance. While using domain-specific languages designed purely for automated execution \cite{xu_core_2024} might offer theoretical advantages, they would create significant barriers for human practitioners and complicate long-term maintenance.

\subsection{Execution DAG Extraction}\label{subsec:execution_flow}

\begin{figure}[ht]
    \centering
    \includegraphics[width=.85\linewidth]{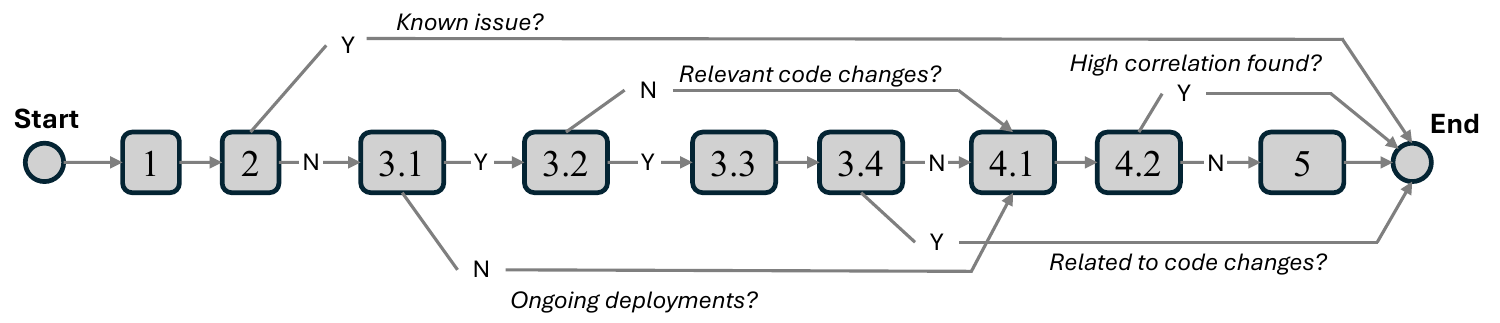}
    \caption{The execution DAG of the example TSG shown in Fig. \ref{fig:example_tsg}. The conditional edges are associated with the key questions and labels: ``Y'' or ``N'', and other edges are unconditional.}
    \label{fig:example_dag}
\end{figure}

We represent the execution DAG as a directed acyclic graph $G=(V, E)$ that captures the execution flow of TSG steps. Each node $v \in V$ corresponds to a TSG step (or sub-step), while each edge $e \in E$ represents a transition between two steps, named as ``edge\_stepX\_stepY''. Edges can be either conditional (dependent on specific outcomes) or unconditional (always executed). 
Extracting the DAG from a TSG provides two key benefits: first, it transforms the ambiguous natural language of the TSG into a formalized workflow; second, it enables a formal representation of step dependencies.
Then, the scheduler can determine the next step node(s) and when to trigger the execution of a node on the DAG based on the states of its incoming edges. This process will be discussed in Section \ref{subsubsec:scheduler}. 
Figure \ref{fig:example_dag} visualizes the execution DAG for our example TSG from Fig. \ref{fig:example_tsg}. We added two special nodes, \textit{Start} and \textit{End}, to mark the beginning and end of the execution process.

We implement execution DAG extraction using an LLM-based approach that automatically extracts execution DAGs from TSG documentation. Our method employs a specialized prompt to analyze TSG structure and generate the DAG in a standardized JSON format. The extraction process systematically identifies individual steps and their interconnections, mapping each TSG step to a DAG node while preserving the original step numbering. For each node, we extract three key components: (1) step number, (2) a one-sentence description, and (3) output edges with their associated conditions. 
Although today's LLMs can extract DAGs with high precision, as we will evaluate in Section \ref{subsec:rq1}, we still involve human experts—the TSG authors—to review the results.

Representing TSG workflows as DAGs is not a new concept, but our method takes a novel approach. Unlike Nissist~\cite{an_nissist_2024}, which builds a comprehensive knowledge graph, or LLexus~\cite{las-casas_llexus_2024}, which creates executable workflows, our approach generates a lightweight DAG. This DAG focuses solely on the essential dependencies and conditional logic, serving as supplementary metadata to the original documentation rather than a complete replacement.
This specific design offers two major benefits: streamlined maintenance, as the DAGs remain synchronized with any changes to the original TSG, and simplified verification, as it is far easier to ensure the DAG's correctness than with the more complex, self-contained graphs used in other approaches.

\subsection{Query Preparation Plugin (QPP) Extraction}
\label{subsec:plugin_extraction}
\begin{wrapfigure}{r}{0.45\textwidth}
\centering
\includegraphics[width=\linewidth]{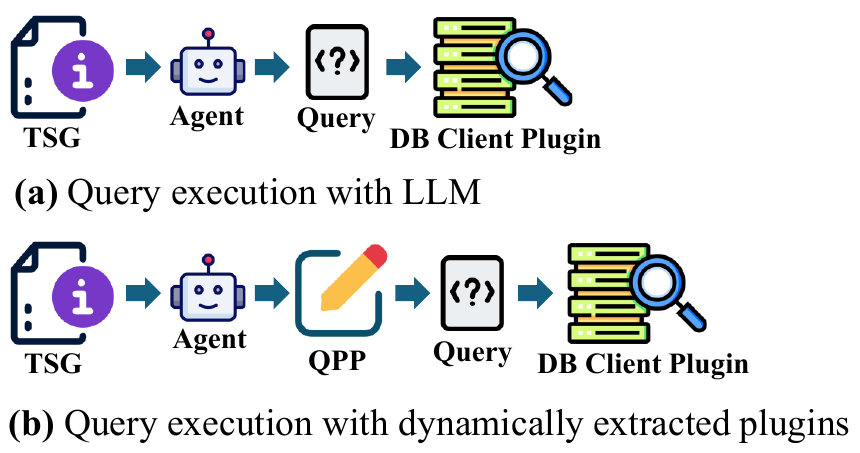}
\caption{The process of running a log query.}
\label{fig:kusto_query}
\end{wrapfigure}

The conventional approach for an LLM-based agent involves generating a query from the TSG's instructions and then invoking the log database client using a plugin, as depicted in Fig. \ref{fig:kusto_query} (a). This text-to-query generation is a challenging task \cite{liu2025nl2sql, liu2025surveynl2sql, tang2024nl2kql}, even with query templates provided in the TSGs. The query templates are often lengthy and complex, and because the KQL is less prevalent than languages like SQL or Python, today's LLMs are less proficient at generating it. Errors like misinterpreting a template or using incorrect parameters not only cause the query to fail but also trigger a costly and time-consuming correction loop for the agent, wasting both time and tokens.

A critical observation is that these queries are structured as templates, utilizing either explicit placeholders (e.g., ``where DeployRing == `\{ring\}' '') or implicit ones that could be inferred from context (e.g., ``replace the StartTime to the incident's start time when using the query''). Leveraging this characteristic, we propose using an LLM to extract these templates and generate dedicated plugins for automated query preparation.

A QPP is responsible for composing a complete query from provided parameters, which the database client can then execute. Inside the QPP, the placeholders will be replaced with the parameters. For example, a call to ``QPP1(..., ring = `test',...)'' results in a query containing ``...where DeployRing == `test'...''.
This new workflow, shown in Fig. \ref{fig:kusto_query} (b), represents a significant improvement over the previous approach. The agent now simply calls the specialized QPP with the required parameters to get a valid query. This query is then passed directly to the client plugin, which not only reduces generation errors but also greatly improves the overall efficiency of TSG execution.
As part of the preprocessing stage, we use an LLM to automatically extract QPPs from TSG documentation. We shall evaluate its accuracy in Sec. \ref{subsec:rq1}.

\subsection{DAG-Guided TSG Execution}
\label{subsec:agent}
In this section, we discuss how to automate TSG execution in \method. 
As illustrated in Fig. \ref{fig:approach}, our agent architecture consists of four core components: (1) a Scheduler, (2) a group of Executors, (3) a memory system, and (4) a set of plugins.
When an incident is triggered, the agent is activated with the incident ID. The Scheduler first locates the corresponding TSG, its execution DAG, and the QPPs, which are all stored together. Guided by the DAG, the Scheduler then begins a job scheduling the Executors to work on the steps. When the end node is reached, the Scheduler concludes the process and presents the final conclusion to the user. We will now discuss the core components.

\subsubsection{Scheduler}
\label{subsubsec:scheduler}
Our scheduler design follows an event-driven model similar to prior workflow systems~\cite{welsh2001seda, isard2007dryad, dean2008mapreduce}, but tailored to TSG execution.
The Scheduler initializes TSG execution by adding the start node to a ready-to-execute queue. When an Executor is available, it is assigned a node from this queue. The Scheduler then operates in an event-driven manner, responding to events like step completion or failure. 
It updates node and edge states to determine subsequent executable steps. Each DAG node and edge maintains one of three states: \textit{unknown}, \textit{enabled}, or \textit{disabled}. Initially, all elements are unknown except for the start node, which is enabled. Upon successful node completion, outgoing edges become enabled if no condition exists or if the condition is satisfied; otherwise, they are disabled. A node becomes enabled when all incoming edges are non-unknown and at least one is enabled.
The Scheduler terminates upon reaching the end node or when no enabled nodes remain in the queue.

TSG step execution may fail due to plugin failures or step errors. The Scheduler implements retry mechanisms with configurable limits before marking steps as failed, and failed nodes disable all outgoing edges to prevent downstream execution.

\subsubsection{Executor}
\label{subsubsec:executor}
An Executor is an agent responsible for executing individual TSG steps. The Scheduler provisions an Executor when a step is ready to run, initializing its context with the necessary information: incident report, TSG documentation, plugins, the corresponding DAG node/edges of the current step, and the execution history. This ensures that an Executor is aware of what to do for the current step as well as all previously completed steps. 

The Executor is explicitly restricted to a single, specified TSG step. It uses an iterative reasoning-action cycle \cite{yao2023react} with chain-of-thought (CoT) reasoning \cite{wei2022cot, wang2024cot} to understand and perform the sub-task. This process continues until the step either succeeds or fails. Upon successful completion, the Executor summarizes the results and, based on the step's outcome, updates the state of its outgoing edges to either enabled or disabled. For example, as shown in the execution DAG in Fig. \ref{fig:example_dag}, the Executor for Step 3.1 will enable ``edge\_step3.1\_step4.1'' and disable ``edge\_step3.1\_step3.2'' if no ongoing deployments right before the incident.

This schedule-executor design naturally fits the step-by-step nature of TSGs, as each Executor reasons, observes, and acts on the natural language instructions of its assigned sub-task.
While our design falls into the multi-agent paradigm common in the literature \cite{langgraph, wu2024autogen, chaoyun2025allhands, hong2024metagpt, autogpt, taskweaver}, it adopts a different philosophy. Most existing work advocates for collaboration between specialized agents, whereas ours focuses on scheduling homogeneous executors.

\subsubsection{Plugins}
\label{subsubsec:plugins}

We created a suite of plugins to support all tasks within TSGs, building on our analysis of tool usage in Section \ref{subsec:tsg_scope}. This suite includes:  i) a log data retrieval plugin that takes a KQL query and returns the log data; ii) a DevOps plugin for tasks like checking deployments and getting code changes; iii) a metric data retrieval plugin for time-series data based on the metric name and a time frame. iv) a code interpreter for generating and running Python code. 
In addition to these general-purpose tools, we introduced Query Preparation Plugins (QPPs) in Sec. \ref{subsec:plugin_extraction}. These are a special type of plugins for specific TSGs. Therefore, when a TSG is loaded, the plugin pool is composed of both the general-purpose plugins and the QPPs associated with that specific TSG.

\subsubsection{Memory}
\label{subsubsec:memory}

\begin{wrapfigure}{r}{0.45\textwidth}
\centering
\includegraphics[width=\linewidth]{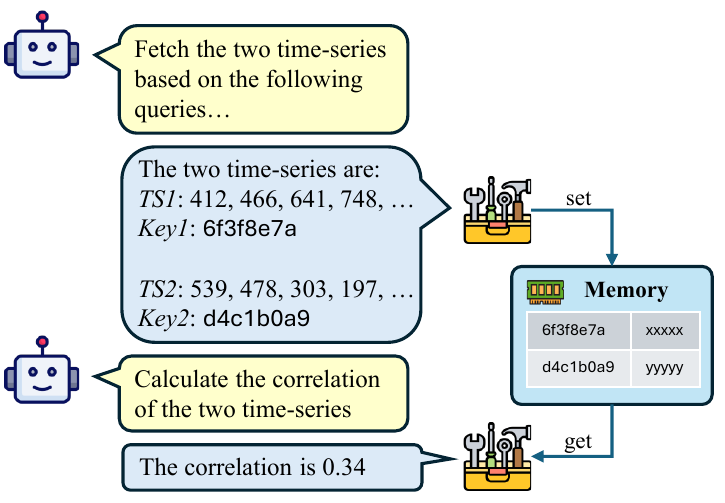}
\caption{Cross-plugin data flow via memory.}
\label{fig:correlation}
\end{wrapfigure}

Completing tasks in a TSG often requires using multiple tools in a complementary way.
For a concrete example, in the TSG from Fig. \ref{fig:example_tsg}, Step 4.1 requires pulling two time-series datasets, and Step 4.2 then calculates the Pearson correlation between them. This process uses two plugins: the metric data retrieval plugin and the code interpreter.
Without a dedicated memory system, the data retrieval plugin would return the time series data as plain text, cluttering the agent's conversation history. The code interpreter would then have to generate code to parse these large lists of numbers and assign them to variables before performing any calculations, a process that is both time- and token-consuming and error-prone.
By using a memory system, the agent's workflow is much more elegant. The metric data retrieval plugin gets the time series data and stores it directly in memory. The code interpreter then accesses these datasets via simple memory references to generate and execute the correlation code. This process returns a quantitative result without the need to pass large data payloads as text messages. This end-to-end process is visualized in Fig. \ref{fig:correlation}, where the data retrieval plugin returns only sample data and the memory reference keys to the agent.

The memory architecture follows a key-value paradigm where string-based keys provide unique data identification and values accommodate arbitrary data types, ranging from primitive types (strings, numbers, lists, dictionaries) to complex structured data (DataFrames, ndarrays). 
Our implementation leverages MongoDB~\cite{mongodb} as the underlying storage backend, because it supports these diverse data types and offers a scalable architecture for the large data volumes common in enterprise TSG operations. While our current implementation utilizes MongoDB, the memory interface abstracts storage implementation details, enabling seamless migration to alternative database solutions with minimal integration overhead.

\subsection{Parallelized TSG Execution}\label{subsec:parallelization}
So far, we have discussed how \method~ executes TSGs on a step-by-step basis. In other words, there is only one active Executor when \method~ is executing a non-parallelized TSG. However, as discussed in Finding 2 (Section \ref{subsec:tsg_scope}), many TSGs can be parallelized, inspired by how a group of SREs would collaborate. Using the example TSG in Fig. \ref{fig:example_tsg}, if a group of SREs were to collaboratively work on a single incident, they could be assigned to different task streams. For instance, three SREs could concurrently work on Steps 2, 3, and 4, respectively, as these steps are independent of each other. After completing their respective sub-tasks, they can reconvene to conclude the final decision.
Another common pattern in many TSGs involves multiple steps that retrieve different prerequisite data before summarizing the findings. A significant advantage of this collaborative diagnosis is the reduction in time to resolve an incident.

\method~ naturally supports this collaborative paradigm, with the Scheduler capable of allocating multiple Executors to work on independent steps concurrently. However, this approach faces a practical challenge: existing TSGs do not explicitly state the dependencies between their steps. To enable parallel execution, TSG steps must first be parallelized by identifying independent steps and modifying the TSG documentation to explicitly state their independence. This typically involves revising the transitions between steps in an existing TSG, which usually requires modifying only a few sentences.

\begin{figure}[t]
    \centering
    \includegraphics[width=0.85\linewidth]{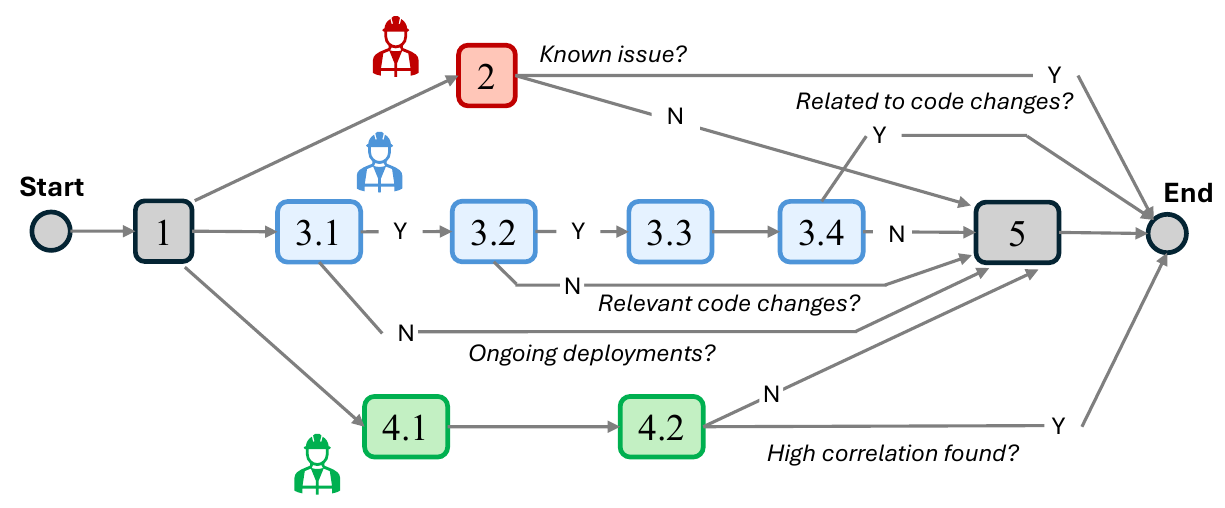}
    \caption{The execution DAG of the TSG in Fig. \ref{fig:example_tsg} after parallelization. The steps that can run concurrently are colored in red, blue, and green, and multiple Executors could be assigned to work on different sub-tasks.}
    \label{fig:parallelized_dag}
\end{figure}

Taking the TSG in Fig. \ref{fig:example_tsg} as an example, its original execution DAG (Fig. \ref{fig:example_dag}) shows that steps are executed sequentially.
Since Step 2, 3, and 4 are independent, we can modify the TSG to allow \method~to allocate multiple Executors to these three sub-tasks. The modifications to the original TSG are minor. For example, the last sentence in Step 1 is changed from ``Now, proceed to Step 2'' to ``Now, start Step 2, Step 3.1, and Step 4.1, simultaneously''. Other modifications are similar, simply redirecting to a different step.
After parallelizing the TSG, a new execution DAG is extracted based on what we have discussed in Sec. \ref{subsec:execution_flow}, as shown in Fig. \ref{fig:parallelized_dag}. Comparing this new DAG to the original (Fig. \ref{fig:example_dag}), two key differences are apparent: 1) Three sub-tasks are initiated after Step 1; 2) The ``fallback'' paths of the three sub-tasks now point to Step 5. This means that if any sub-task reaches a conclusion, the whole process ends; otherwise, it falls back to Step 5 to involve an SRE for further investigation.
Semantically, the parallelized TSG is identical to the original TSG in Fig. \ref{fig:example_tsg}.

While the modifications to an existing TSG are usually minor, they require deep domain knowledge to identify parallelization opportunities. One must understand, for instance, if steps are logically or data-wise dependent. In this work, we collaborated with SREs to identify these opportunities in TSGs. The possibility of using LLMs to automate this process is an intriguing direction, which we leave for future work.

\section{Experimentation}\label{sec:experimentation}
In this section, we will discuss the experimental setup and the evaluation of our proposed approach. 
Our implementation of \method~ consists of 6,698 lines of Python code, excluding the prompts. We implemented Executors as individual processes.

\subsection{Datasets}\label{subsec:datasets}
Our evaluation set consists of 15 representative and widely-used TSGs, drawn from our broader study in Sec. \ref{subsec:tsg_scope}, and 80 real incidents sampled from a large-scale production system at Microsoft. We selected these TSGs to ensure diversity in both incident type and complexity. The final dataset represents typical production scenarios, including service availability degradation, latency issues, pipeline failures, and request volume anomalies. Their complexity varies substantially, ranging from concise 3-step procedures to comprehensive 28-step diagnostic workflows, with a mean complexity of 9 steps per guide. These TSGs were comprehensively revised by experienced SREs following our quality guidelines (Section \ref{subsec:guidelines}) to achieve sufficient clarity, precision, and usability.

\subsection{Research Questions}\label{subsec:research_questions}
Our experimental evaluation investigates the following research questions to assess the effectiveness and practical viability of our proposed TSG automation framework:

\begin{itemize}
    \item \textbf{RQ1}: \textit{Preprocessing Accuracy} -- To what extent can LLM-based preprocessing accurately extract execution DAGs and QPPs from TSG documentation?
    \item \textbf{RQ2}: \textit{System Performance} -- How effectively does \method~ improve TSG execution reliability and efficiency compared to baseline approaches, and what are the key performance drivers?
    \item \textbf{RQ3}: \textit{Parallelization Benefits} -- What performance gains can be achieved through parallelization of independent TSG steps?
\end{itemize}

\subsection{RQ1: Preprocessing Accuracy}\label{subsec:rq1}

\paragraph{Execution DAG Extraction}
We evaluated our DAG extraction accuracy by comparing extracted DAGs with manually annotated ground truth, which were initially generated by an LLM and then corrected by experienced SREs.
Our approach was assessed at three levels: \textit{node-level} metrics for step identification, \textit{edge-level} metrics for dependency correctness, and \textit{condition-level} metrics for the semantic alignment of conditional expressions. For node- and edge-level metrics, we computed precision, recall, and F1-score. A false positive is an extracted element not in the ground truth, while a false negative is a ground truth element that was not extracted. For conditional expressions, we leveraged an LLM to evaluate semantic alignment.

We conducted three independent experiments across the 15 TSGs to ensure statistical robustness. As presented in Table~\ref{tab:plandag_results}, our experimental results demonstrate consistently superior extraction performance with minimal variance across experimental runs. The aggregate F1-score achieves $\sim$94.89\%, indicating robust overall performance. Detailed analysis reveals that node identification attains near-optimal performance (exceeding 99\% across all evaluation metrics), edge extraction demonstrates strong performance characteristics ($\sim$93.47\% F1-score), and condition semantic matching achieves $\sim$91.78\% accuracy. These empirical findings substantiate that LLM-based execution DAG extraction delivers satisfactory performance for automated troubleshooting workflows, requiring minimal manual intervention to rectify extracted DAG representations.

\begin{table}[t]
\centering
\small
\caption{Execution DAG Extraction Results (Mean ± Std across 3 experiments)}
\label{tab:plandag_results}
\begin{tabular}{lcccc}
\toprule
\textbf{Level} & \textbf{Precision (\%)} & \textbf{Recall (\%)} & \textbf{F1-Score (\%)} & \textbf{Accuracy (\%)} \\
\midrule
Overall & 94.20 ± 0.91 & 95.73 ± 0.43 & 94.89 ± 0.67 & -- \\
Node-Level & 99.63 ± 0.65 & 99.24 ± 0.03 & 99.42 ± 0.33 & -- \\
Edge-Level & 91.19 ± 0.92 & 96.15 ± 0.05 & 93.47 ± 0.50 & -- \\
Condition-Level & -- & -- & -- & 91.78 ± 1.23 \\
\bottomrule
\end{tabular}
\end{table}

Our study of the failed cases revealed that edge-level errors result from incorrectly inferred dependencies, such as the addition of spurious edges to the graph. Condition-level errors primarily occur due to a lack of precision in the conditional expressions, often manifested as an oversimplification. In practice, however, such oversimplified expressions may not cause real issues, as the DAG is used complementarily with the original TSG documentation.


\paragraph{QPP Extraction}
Our evaluation on QPP extraction follows the same methodology as DAG extraction.
The 15 TSGs selected in Section \ref{subsec:datasets} contain a total of 86 query templates. 
We compare the extracted plugins against manually annotated ground truth. The annotation process involves identifying the query template and the parameters in the TSG documentation. The labeled templates are then used as the ground truth for evaluation.

We don't evaluate the extracted templates based on an exact match with the ground truth, but rather on their functional equivalence. This is because an LLM may generate a template with different formatting, such as varied whitespace or newlines, or use different placeholder names (e.g., ``user\_id'' vs. ``UserId'') while the underlying logic remains identical. To account for these minor variations, we use a diff tool to perform a line-by-line comparison. We consider the extraction successful if manually reviewing the diff results confirms the template is functionally equivalent to the ground truth.

To ensure the robustness and consistency of our results, we ran the extraction process three times for each TSG, resulting in a total of 258 extractions. Our LLM-based approach achieved a success rate of 97.3\%, with 251 successful extractions. All failed extractions were due to the LLM making mistakes with escape characters, such as adding an extra backslash, which caused the extraction to fail.
This experiment validates that LLM-based QPP extraction is highly accurate and requires only minimal human intervention.

\subsection{RQ2: System Performance}\label{subsec:rq2}
In this section, we evaluate the performance of our proposed TSG automation framework by comparing it against baseline approaches.
We selected ReAct and TaskWeaver as baselines because they are publicly available and have been used for comparison in prior work~\cite{zhang2024flash}. 

We compare \method~ against two baseline approaches, all sharing the same system prompt and plugin set as \method, except for the QPPs. 
We evaluated all methods on a common set of TSGs, which were revised according to our guidelines in Sec.~\ref{subsec:guidelines}, to guarantee a fair comparison on an identical documentation baseline.
\begin{itemize}
    \item \textbf{ReAct}: We implemented a baseline based on the reasoning and action paradigm~\cite{yao2023react}. Specifically, the agent operates in a loop of ``thought-action-observation'' to iteratively reason about the user request, decide on actions, and observe their results. For each iteration, we adopt Chain-of-Thought (CoT)~\cite{wei2022cot} prompting to enhance reasoning capabilities. ReAct has no pre-computed plan, relying entirely on online reasoning to navigate through TSG steps.
    \item \textbf{TaskWeaver}: We implemented a baseline using TaskWeaver~\cite{taskweaver}, which is a state-of-the-art multi-agent framework for data-intensive tasks. TaskWeaver employs a dual-agent architecture with a Planner agent that decomposes the user request into subtasks and a Code Interpreter agent to generate and execute Python code for data processing.
\end{itemize}
 
In addition to the baselines, we also evaluate the performance of \method~ with and without the QPPs. This allows us to assess the impact of QPP extraction on TSG execution performance. The two variants are referred to as \textbf{\method}~ and \textbf{\method~ w/o QPP}, respectively.

We conducted our experiments using four different LLMs to evaluate a range of model capabilities. We chose GPT-4.1, which represents the latest generation of LLM at the time of writing, along with GPT-4o and Grok-3, two widely adopted high-performance models. GPT-4.1-mini was included as a representative of a moderate-performance LLM.

To assess the effectiveness and performance characteristics of our proposed framework, we employ the following metrics:
\begin{itemize}
    \item \textbf{Success Rate} Success rate quantifies the percentage of successful TSG executions. \textit{A TSG execution is classified as successful when both the agent's execution path and the diagnostic conclusion align with the ground truth human label}. 
    \item \textbf{Execution Latency} Execution latency measures the temporal duration required for complete TSG processing, encompassing the interval from initial TSG invocation to final execution termination. 
    \item \textbf{Total Token Consumption} Total token consumption quantifies the cost through aggregate input and output token utilization during TSG execution.
\end{itemize}

The execution path is the sequence of steps the agent executes before reaching a conclusion. 
Since we used past incidents for evaluation, we already knew the ground truth diagnostic conclusion for each incident. 
Latency and token consumption metrics are computed only for successful TSG executions to ensure meaningful performance comparisons.

The success rates under different LLMs are shown in Table \ref{tab:success_rates}.
\method~ outperforms all baselines across different LLMs, achieving its highest success rate of 94.38\% under GPT-4.1. Our framework showed a maximum gain of $\sim$20\% over TaskWeaver and React with GPT-4o, and a minimum gain of $\sim$8\% with Grok-3. Even when the QPPs are removed, as in the StepFly-wo-QPP variant, our framework still outperforms both baselines. Comparing \method{} with StepFly-wo-QPP demonstrates the consistent value of QPP extraction: improvements range from 2.5\% (GPT-4.1) to 10.63\% (GPT-4.1-mini). The benefit is more pronounced with less capable LLMs, as they are more prone to query generation errors---a primary source of failures in TSG execution. Even for powerful LLMs like GPT-4.1 and Grok-3, QPP still provides meaningful gains by eliminating on-the-fly query generation risks. This finding is consistent with our earlier evaluations in Sec. \ref{subsec:rq1}, which showed that powerful LLMs can extract the query templates with near-perfect accuracy.

\begin{table}[t]
\centering
\caption{Execution Success Rates (Mean ± Std across 3 experiments)}
\label{tab:success_rates}
\begin{tabular}{lllll}
\toprule
Model & GPT-4.1 & GPT-4.1-mini & GPT-4o & Grok-3 \\
Method & & & & \\
\midrule
\textbf{React} & 71.25 $\pm$ 5.3 & 61.25 $\pm$ 3.54 & 66.88 $\pm$ 2.65 & 72.5 $\pm$ 1.77 \\
\textbf{TaskWeaver} & 80.0 $\pm$ 1.77 & 68.12 $\pm$ 4.42 & 72.5 $\pm$ 3.54 & 80.62 $\pm$ 4.42 \\
\textbf{StepFly-wo-QPP} & 91.88 $\pm$ 2.65 & 73.75 $\pm$ 7.07 & 81.88 $\pm$ 0.88 & 85.62 $\pm$ 4.42 \\
\textbf{StepFly} & \textbf{94.38} $\pm$ 0.88 & \textbf{84.38} $\pm$ 2.65 & \textbf{92.5} $\pm$ 1.77 & \textbf{88.75} $\pm$ 0.63 \\
\bottomrule
\end{tabular}
\end{table}


Our analysis showed that React performed the worst across all LLMs. We found its most frequent failures were stopping mid-process and generating incorrect queries. For example, React, the single-agent design, would often express the intention to proceed but fail to invoke the necessary tools, breaking the reasoning-action loop. While TaskWeaver's dual-agent design mitigated this, its primary issue was executing incorrect steps. Both of these problems are mitigated in \method~ because our framework introduces an execution DAG that explicitly controls the workflow and ensures the agent strictly adheres to the defined path.

\begin{wrapfigure}{l}{0.6\textwidth}
    \centering
    \includegraphics[width=\linewidth]{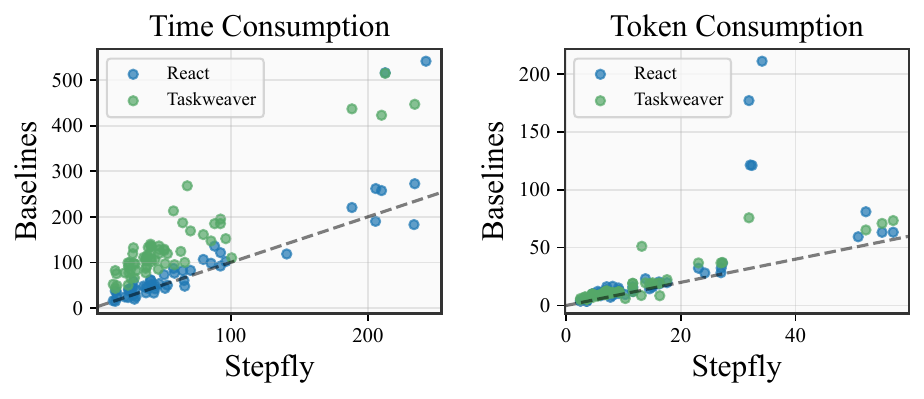}
    \caption{Comparison of time (left) and token consumption (right) between \method~ and baselines under GPT-4.1. The absolute values are scaled for comparison only.}
    \label{fig:consumptions}
\end{wrapfigure}

Figure \ref{fig:consumptions} provides a comparison between \method~ and the two baselines in terms of time and total token consumption under GPT-4.1. We conducted the comparisons under different models, which yield similar results.
Figure \ref{fig:consumptions} (left) compares the time consumption of \method~ with baselines. The dashed line represents equal time consumption; points above it indicate that the baseline is slower than \method. \method~ and React show similar time consumption, with the median time cost of \method~  being 88\% that of React. TaskWeaver's dual-agent design leads to increased back-and-forth communication, which results in a much higher time overhead. The median time consumption of \method~ is only 38\% of TaskWeaver. 

The observation on token consumption is similar to the time consumption analysis, as shown in Fig. \ref{fig:consumptions} (right). Our framework, \method, demonstrates superior efficiency, with its median token consumption being only $\sim$71\% of React's and $\sim$61\% of TaskWeaver's. The notable outliers in React's token consumption are due to its lack of a structured memory system. It's forced to present all data as raw text within the prompt. In incidents where a plugin returns a large volume of data, this approach leads to a significant increase in prompt size and, consequently, token consumption.



\subsection{RQ3: Parallelization Benefits}\label{subsec:rq3}

In this section, we evaluate the benefits of parallelizing TSGs to determine if it saves time during diagnosis, which is essential for timely incident management. We identified 7 of the 15 TSGs from our study as having potential for parallelization. After the TSGs being revised according to Sec. \ref{subsec:parallelization}, we extracted the execution DAGs. For our evaluation, we ran the TSGs with varying numbers of available Executors, ranging from 2 to 5. The number of Executors determines the maximum number of triggered nodes on the execution DAG that can be run concurrently. To ensure the robustness of our results, we ran three independent experiments for each incident, measuring the end-to-end time consumption.

\begin{figure}[t]
    \centering
    \includegraphics[width=0.7\linewidth]{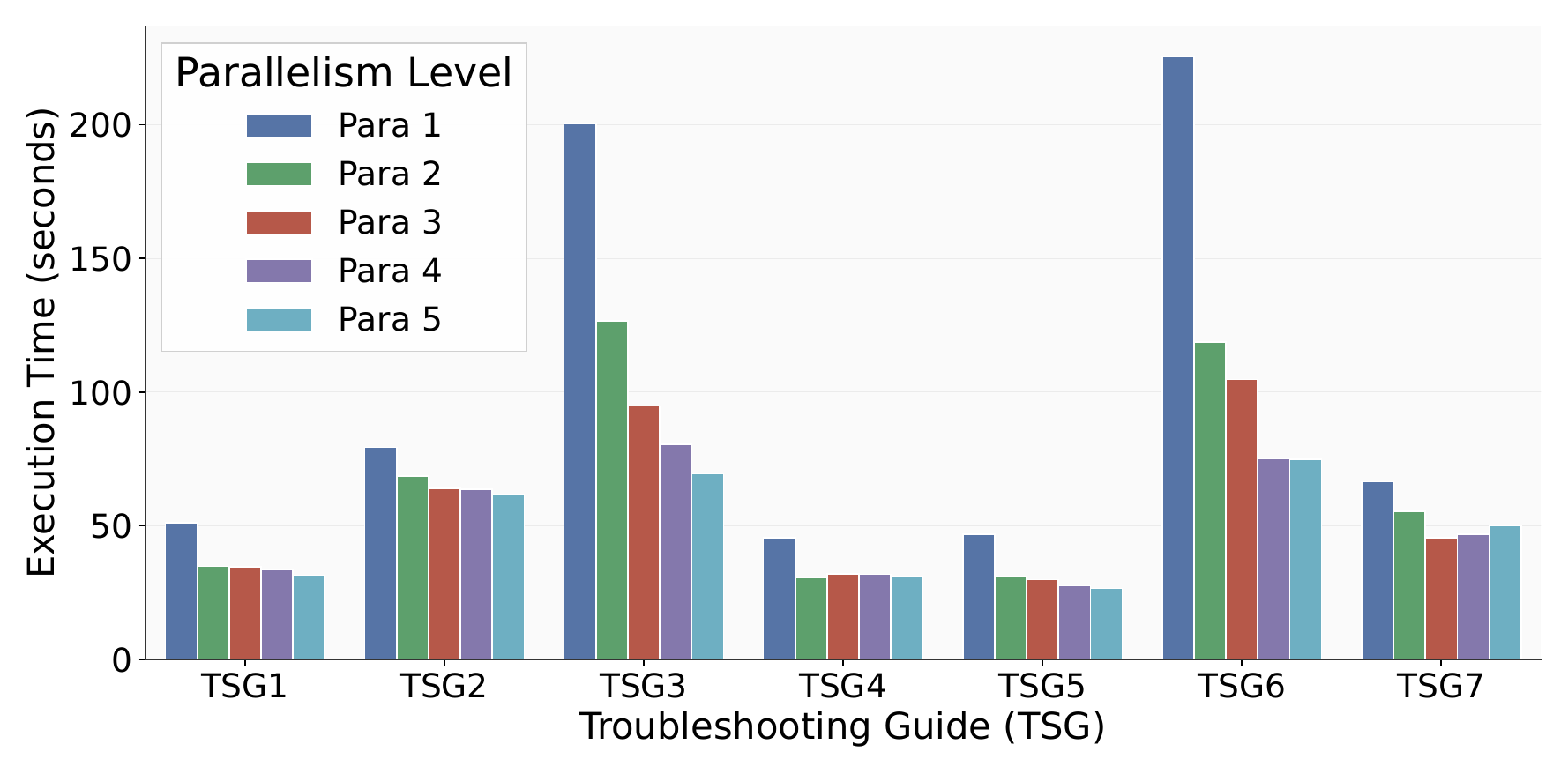}
    \caption{Time consumption comparison between sequential execution and parallelized execution with different numbers of Executors.}
    \label{fig:rq3_time_usage}
    \vspace{-5pt}
\end{figure}

As shown in Fig. \ref{fig:rq3_time_usage}, parallelization successfully reduces the time consumption for all evaluated TSGs compared to sequential execution. We observe that as the number of Executors increases, the execution time either steadily decreases or reaches a saturation point with minor fluctuations. In fact, increasing the number of available Executors beyond the maximum degree of parallelism provides no further benefit. For example, as seen in Fig. \ref{fig:parallelized_dag} for our example TSG, using more than 3 Executors does not reduce time as the maximum degree of parallelism is 3.

The most significant improvements were observed in TSGs with more independent steps that can run concurrently. For instance, TSG6 showed a 70.6\% reduction in execution time, improving from 243.36 seconds (sequential) to 71.57 seconds (with 5 Executors). Similarly, TSG3 achieved a 70.4\% improvement, reducing its time from 207.93 seconds to 61.59 seconds. Less gain was observed in other guides, with TSG7 showing a reduction of 32.9\%. TSG2 in Fig. \ref{fig:rq3_time_usage} shows the execution time reduction for the parallelized version of our example TSG in Fig. \ref{fig:example_tsg}. Overall, the mean reduction with 5 Executors across all TSGs was 51.31\%.

\subsection{Case Study}\label{subsec:case_study}

We present two representative TSGs to illustrate the practical benefits of \method \ beyond the quantitative results shown in RQ2 and RQ3.

\subsubsection{Case 1: Parallelization for Faster Root Cause Identification}\label{subsec:case1}

We've been using the example TSG in Fig. \ref{fig:example_tsg} (TSG2 in Fig. \ref{fig:rq3_time_usage}) across this paper, which is used for diagnosing an availability issue for a large-scale service. To recap, the TSG contains 5 major steps, where Steps 2-4 execute independent diagnostic checks in parallel, as shown in Fig. \ref{fig:example_dag}. Our analysis of incidents associated with this TSG found that around 72.7\% of cases ultimately identified the root cause in Step 4.2—that is, the issue was caused by a dependent service. However, when this TSG is followed in a sequential manner, reaching Step 4.2 requires completing all preceding steps, resulting in significant delays.

The parallelized TSG enables Steps 2-4 to execute concurrently after Step 1. This design is particularly effective because Step 4 can begin immediately without needing to wait for Steps 2 and 3, which are used to check for less-common issues like known bugs and deployment problems. In several real incidents in our experiments in Sec. \ref{subsec:rq3}, Step 4.2 identified high correlations while Step 3 was still analyzing deployment data, allowing it to be terminated early. This early discovery enabled the immediate termination of unfinished investigations and ticket transfer, avoiding the execution of remaining steps that would have been required in sequential processing. 

\subsubsection{Case 2: QPPs and Memory System Efficiency}\label{subsec:case2}

We can use TSG6 from Fig.\ref{fig:rq3_time_usage} to demonstrate how QPPs and the memory system (discussed in Sec. \ref{subsubsec:memory}) can efficiently handle large data volumes. This TSG performs a comprehensive reliability analysis through 11 steps, many of which require executing complex queries that return large datasets. 

Taking one real incident as an example, without QPPs, an agent would need to generate 9 distinct queries on the fly, totaling 127 lines of KQL. These queries range from 11 lines (shortest) to 27 lines (longest) per template, with intricate syntax that is highly prone to generation errors. Using QPPs, each query execution requires only parameter specification (typically 5-10 parameters) rather than full query generation. 
Furthermore, our memory system efficiently handles the large data volumes returned from these executed queries. For instance, the top 3 queries returned datasets of 394, 151, and 759 rows. With our memory system, the 394-row structured data (approximately 25KB) was stored as a DataFrame and represented in the context as a description of its schema and a 3-row sample (less than 200 tokens).

The cumulative effect across all 9 query executions was substantial: while the actual queries totaled 127 lines and returned data exceeded 50KB, the context representation required less than 2KB. This 96\% reduction not only prevented context window overflow but also eliminated the token cost and error risk of generating complex KQL queries that LLMs struggle with.

\section{Threats to Validity}\label{sec:threat}

First, the construction and parsing of TSGs rely on heuristics and templates derived from a limited corpus of existing TSGs and our interactions with domain experts. 
While the TSG Mentor tool helps normalize documentation, the inherent variability in TSG authoring across organizations and domains may limit the generalizability of our documentation enhancement pipeline. We mitigated this by iteratively refining our heuristics with feedback from multiple service teams and including diverse formats in our evaluation.

Second, the agentic execution framework depends on LLM-generated plans and decisions, which are inherently non-deterministic. This can impact the reproducibility and consistency of TSG execution across different runs or prompt formulations. To address this, we mitigated the threat by fixing model versions and temperature settings during our evaluation. We also guided agent decisions along known resolution paths using our execution DAGs.

Finally, our evaluation introduces a potential threat to external validity because it was conducted on a select set of frequently used TSGs within a controlled scope of service domains. This means our results may not generalize to TSGs with highly specialized procedures, external system dependencies, or those requiring human judgment beyond current LLM capabilities. To address this, future work should explore broader deployment across varied operational domains and include more heterogeneous TSGs to fully assess robustness.

\section{Discussion}\label{sec:discussion}

\paragraph{Practicality}
A prototype of \method~ is deployed where it is automatically triggered by incoming incidents. The agent retrieves the relevant TSG and initiates step execution. To provide transparency and traceability, the entire execution process, including queries and other actions, is logged in a structured format and posted to a dedicated channel in the incident management system. This allows SREs to review the agent's actions and conclusions, ensuring that human oversight is maintained while automating routine tasks. The prototype's weekly usage, triggered hundreds of times by over 170 monitors from more than 70 teams, has reduced the median incident mitigation time by $\sim$34\%, demonstrating its practical utility in real-world operations.

\paragraph{Extensibility}
\method~ is architected with extensibility principles, which gives it broad applicability across DevOps and IT operations domains. Its fundamental capability as a workflow-aware agent allows it to interpret and autonomously execute documented procedures. The framework's modular design simplifies integration with new systems (e.g., network appliances, database clusters), as it primarily involves extending the action library with domain-specific commands, leaving the core planner and executor components unchanged. Additionally, \method~ supports plugin mechanisms for external tools and services, enabling developers to define custom actions. 
\method~ offers a scalable foundation for automating operational procedures across diverse technical domains.

\paragraph{Safety} The \method{} framework operates with human oversight, where SREs review the agent's conclusions before taking any mitigation actions. The current implementation focuses primarily on diagnostic operations, which are read-only in nature---querying logs, retrieving metrics, and analyzing data---rather than executing potentially destructive remediation actions. All agent actions, including queries executed and decisions made, are comprehensively logged and posted to the incident management system, ensuring full transparency and traceability. This design allows SREs to audit the agent's reasoning process and intervene when necessary.

\paragraph{Token Consumption of Parallelized Execution}
While we've shown that parallelizing TSG execution can significantly reduce time, its impact on token consumption is more complex. For instance, as shown in Fig. \ref{fig:example_dag}, if a known issue is found in Step 2, a sequential execution would stop there, saving tokens. However, the parallelized version shown in Fig. \ref{fig:parallelized_dag} would trigger Steps 2, 3, and 4 simultaneously, potentially wasting tokens. Conversely, if the root cause is a dependent service issue identified in Step 4.2 (as discussed in Sec. \ref{subsec:case1}), the parallelized execution can save tokens by enabling the early termination of Steps 2 and 3. Therefore, the overall token consumption depends on the distribution of incidents' diagnostic outcomes.

\section{Related Work}\label{sec:related}

\paragraph{LLM-powered Incident Management}
Incident management has been a subject of extensive research, with recent work leveraging LLMs to automate tasks such as incident understanding, triage, and root cause analysis \cite{pengxiang2023summary, zexin2024comet, wang2021Groot, xie2024cloudatlas, changhua2025sop}.
TSGs are also a focus of research. Prior work such as DeepRmd \cite{jiajun2020tsg} has focused on recommending TSGs based on incident context, while Nissist \cite{an_nissist_2024} transforms TSGs into knowledge graphs to provide step-by-step guidance.
Automating TSG execution has been explored in prior works, such as AutoTSG \cite{manish2022autotsg}, Flash \cite{zhang2024flash, roy2024rca}, and LLexus \cite{las-casas_llexus_2024}. AutoTSG was an early attempt to automate identifying and executing queries and commands in TSGs. Flash introduces a React-based LLM agent for TSG automation, with a key contribution of dynamically injecting context during execution. LLexus uses LLMs to turn a TSG into a self-contained executable workflow. However, these works do not adequately address the challenges of TSG quality and execution efficiency, which are the primary focus of our work.

\paragraph{LLM-powered Agent Frameworks}
Agent frameworks have emerged to perform multi-step tasks through tool use and environment interaction. Systems like ReAct~\cite{yao2023react}, AutoGPT~\cite{autogpt}, and Toolformer~\cite{schick2023toolformer} showcase how LLMs can reason, plan, and act in open-ended scenarios. However, these frameworks do not specifically address the procedural, state-dependent nature of Troubleshooting Guides (TSGs) and often lack mechanisms for handling data-intensive tasks. TaskWeaver~\cite{taskweaver} and CodeAct~\cite{wang_executable_2024} adopt a code-first approach, where data is stored in the memory of the Python process rather than in the LLM's conversation history. We have implemented a baseline based on TaskWeaver to compare against our proposed framework.

\section{Conclusion and Future Work}\label{sec:conclusion}

This study explored the potential of LLMs for automating TSG execution in incident management. We conducted an empirical study of 92 real-world TSGs, which revealed key findings on their characteristics and quality issues. Guided by these findings, we proposed \method, a three-stage framework featuring quality improvement, preprocessing with execution DAG extraction and Query Preparation Plugins, and an online scheduler-executor architecture that supports parallel execution. Our comprehensive evaluations show that \method \ achieved a $\sim$94\% success rate on GPT-4.1 with reduced time and token consumption. For a subset of TSGs, parallelization further reduced time by 32.9\% to 70.4\%. Overall, our results demonstrate the great potential of TSG automation with LLM-powered agents. While our work focuses on execution, future work is needed to address other key TSG automation tasks, including the creation of guides for new issues, their ongoing maintenance, and automated revision for parallelization.

\section{Data Availability}

Code and synthesized incident data are available at \url{https://github.com/microsoft/StepFly}. 

\begin{acks}
We would like to thank the anonymous reviewers for their valuable comments.
\end{acks}

\bibliographystyle{ACM-Reference-Format}
\bibliography{ref}

@article{landis1977measurement,
  title={The measurement of observer agreement for categorical data},
  author={Landis, J Richard and Koch, Gary G},
  journal={Biometrics},
  volume={33},
  number={1},
  pages={159--174},
  year={1977},
  publisher={JSTOR},
  doi={10.2307/2529310}
}

@inproceedings{yuxuan2024xpert,
  author = {Jiang, Yuxuan and Zhang, Chaoyun and He, Shilin and Yang, Zhihao and Ma, Minghua and Qin, Si and Kang, Yu and Dang, Yingnong and Rajmohan, Saravan and Lin, Qingwei and Zhang, Dongmei},
  title = {Xpert: Empowering Incident Management with Query Recommendations via Large Language Models},
  year = {2024},
  isbn = {9798400702174},
  publisher = {Association for Computing Machinery},
  address = {New York, NY, USA},
  url = {https://doi.org/10.1145/3597503.3639081},
  doi = {10.1145/3597503.3639081},
  booktitle = {Proceedings of the IEEE/ACM 46th International Conference on Software Engineering},
  articleno = {92},
  numpages = {13},
  location = {Lisbon, Portugal},
  series = {ICSE '24}
}

@misc{langgraph,
  author       = {{LangChain AI}},
  title        = {{LangGraph}: State Machines for {LLM} Applications},
  year         = {2024},
  howpublished = {\url{https://github.com/langchain-ai/langgraph}},
  note         = {Accessed: 2025-05-30}
}

@misc{autogpt,
  author       = {Richards, Toran Bruce},
  title        = {{AutoGPT}},
  year         = {2023},
  howpublished = {\url{https://github.com/Significant-Gravitas/AutoGPT}},
  note         = {Accessed: 2025-05-30}
}

@misc{mongodb,
  author       = {{MongoDB, Inc.}},
  title        = {{MongoDB}: The World's Leading Modern Database},
  year         = {2025},
  howpublished = {\url{https://www.mongodb.com/}},
  note         = {Accessed: 2025-05-30}
}

@misc{taskweaver,
  title={{TaskWeaver}: A Code-First Agent Framework},
  author={Bo Qiao and Liqun Li and Xu Zhang and Shilin He and Yu Kang and Chaoyun Zhang and Fangkai Yang and Hang Dong and Jue Zhang and Lu Wang and Minghua Ma and Pu Zhao and Si Qin and Xiaoting Qin and Chao Du and Yong Xu and Qingwei Lin and Saravan Rajmohan and Dongmei Zhang},
  year={2024},
  eprint={2311.17541},
  archivePrefix={arXiv},
  primaryClass={cs.AI},
  doi={10.48550/arXiv.2311.17541},
  url={https://arxiv.org/abs/2311.17541}
}

@inproceedings{chaoyun2025allhands,
  author = {Zhang, Chaoyun and Ma, Zicheng and Wu, Yuhao and He, Shilin and Qin, Si and Ma, Minghua and Qin, Xiaoting and Kang, Yu and Liang, Yuyi and Gou, Xiaoyu and Xue, Yajie and Lin, Qingwei and Rajmohan, Saravan and Zhang, Dongmei and Zhang, Qi},
  booktitle = {2025 IEEE 41st International Conference on Data Engineering (ICDE)},
  title = {{AllHands}: Ask Me Anything on Large-Scale Verbatim Feedback via Large Language Models},
  year = {2025},
  pages = {43--57},
  doi = {10.1109/ICDE65448.2025.00011},
  publisher = {IEEE Computer Society},
  address = {Los Alamitos, CA, USA}
}

@inproceedings{wang2024cot,
  title={Chain-of-Thought Reasoning Without Prompting},
  author={Xuezhi Wang and Denny Zhou},
  booktitle={Advances in Neural Information Processing Systems 37 (NeurIPS)},
  year={2024}
}

@misc{kusto_kql,
  author       = {{Microsoft}},
  title        = {{Kusto Query Language Overview}},
  year         = {2025},
  howpublished = {\url{https://learn.microsoft.com/en-us/kusto/query/}},
  note         = {Accessed: 2026-04-09}
}

@inproceedings{wei2022cot,
  title={Chain-of-Thought Prompting Elicits Reasoning in Large Language Models},
  author={Jason Wei and Xuezhi Wang and Dale Schuurmans and Maarten Bosma and Brian Ichter and Fei Xia and Ed Chi and Quoc Le and Denny Zhou},
  booktitle={Advances in Neural Information Processing Systems 35 (NeurIPS)},
  year={2022}
}

@inproceedings{yao2023react,
  title = {{ReAct}: Synergizing Reasoning and Acting in Language Models},
  author = {Yao, Shunyu and Zhao, Jeffrey and Yu, Dian and Du, Nan and Shafran, Izhak and Narasimhan, Karthik and Cao, Yuan},
  booktitle = {International Conference on Learning Representations (ICLR)},
  year = {2023},
  url = {https://openreview.net/forum?id=WE_vluYUL-X}
}

@inproceedings{hong2024metagpt,
  title={{MetaGPT}: Meta Programming for A Multi-Agent Collaborative Framework},
  author={Sirui Hong and Mingchen Zhuge and Jonathan Chen and Xiawu Zheng and Yuheng Cheng and Jinlin Wang and Ceyao Zhang and Zili Wang and Steven Ka Shing Yau and Zijuan Lin and Liyang Zhou and Chenyu Ran and Lingfeng Xiao and Chenglin Wu and J{\"u}rgen Schmidhuber},
  booktitle={International Conference on Learning Representations (ICLR)},
  year={2024}
}

@article{li2025longcontext,
  title={Long-context {LLMs} Struggle with Long In-context Learning},
  author={Li, Tianle and Zhang, Ge and Do, Quy Duc and Yue, Xiang and Chen, Wenhu},
  journal={Transactions on Machine Learning Research},
  year={2025}
}

@inproceedings{wang2021Groot,
  author={Wang, Hanzhang and Wu, Zhengkai and Jiang, Huai and Huang, Yichao and Wang, Jiamu and Kopru, Selcuk and Xie, Tao},
  booktitle={2021 36th IEEE/ACM International Conference on Automated Software Engineering (ASE)},
  title={Groot: An Event-graph-based Approach for Root Cause Analysis in Industrial Settings},
  year={2021},
  doi={10.1109/ASE51524.2021.9678708}
}

@inproceedings{changhua2025sop,
  author = {Pei, Changhua and Wang, Zexin and Liu, Fengrui and Li, Zeyan and Liu, Yang and He, Xiao and Kang, Rong and Zhang, Tieying and Chen, Jianjun and Li, Jianhui and Xie, Gaogang and Pei, Dan},
  title = {Flow-of-Action: {SOP} Enhanced {LLM}-Based Multi-Agent System for Root Cause Analysis},
  year = {2025},
  isbn = {9798400713316},
  publisher = {Association for Computing Machinery},
  address = {New York, NY, USA},
  doi = {10.1145/3701716.3715225},
  booktitle = {Companion Proceedings of the ACM on Web Conference 2025},
  pages = {422--431},
  numpages = {10},
  location = {Sydney NSW, Australia},
  series = {WWW '25}
}

@inproceedings{NEURIPS2024_babilong,
  author = {Kuratov, Yuri and Bulatov, Aydar and Anokhin, Petr and Rodkin, Ivan and Sorokin, Dmitry and Sorokin, Artyom and Burtsev, Mikhail},
  booktitle = {Advances in Neural Information Processing Systems},
  editor = {A. Globerson and L. Mackey and D. Belgrave and A. Fan and U. Paquet and J. Tomczak and C. Zhang},
  pages = {106519--106554},
  publisher = {Curran Associates, Inc.},
  title = {{BABILong}: Testing the Limits of {LLMs} with Long Context Reasoning-in-a-Haystack},
  volume = {37},
  year = {2024},
  doi = {10.5555/3737916.3741297}
}

@inproceedings{chen2024automatic,
  author = {Chen, Yinfang and Xie, Huaibing and Ma, Minghua and Kang, Yu and Gao, Xin and Shi, Liu and Cao, Yunjie and Gao, Xuechao and Fan, Hao and Wen, Ming and Zeng, Jun and Ghosh, Supriyo and Zhang, Xuchao and Lin, Qingwei and Rajmohan, Saravan and Zhang, Dongmei},
  title = {Automatic Root Cause Analysis via Large Language Models for Cloud Incidents},
  booktitle = {Proceedings of the Nineteenth European Conference on Computer Systems},
  series = {EuroSys '24},
  year = {2024},
  publisher = {ACM},
  address = {New York, NY, USA},
  doi = {10.1145/3627703.3629553}
}

@inproceedings{manish2022autotsg,
  author = {Shetty, Manish and Bansal, Chetan and Upadhyayula, Sai Pramod and Radhakrishna, Arjun and Gupta, Anurag},
  title = {{AutoTSG}: learning and synthesis for incident troubleshooting},
  year = {2022},
  isbn = {9781450394130},
  publisher = {Association for Computing Machinery},
  address = {New York, NY, USA},
  url = {https://doi.org/10.1145/3540250.3558958},
  doi = {10.1145/3540250.3558958},
  booktitle = {Proceedings of the 30th ACM Joint European Software Engineering Conference and Symposium on the Foundations of Software Engineering},
  pages = {1477--1488},
  numpages = {12},
  location = {Singapore, Singapore},
  series = {ESEC/FSE 2022}
}

@inproceedings{pengxiang2023summary,
  author = {Jin, Pengxiang and Zhang, Shenglin and Ma, Minghua and Li, Haozhe and Kang, Yu and Li, Liqun and Liu, Yudong and Qiao, Bo and Zhang, Chaoyun and Zhao, Pu and He, Shilin and Sarro, Federica and Dang, Yingnong and Rajmohan, Saravan and Lin, Qingwei and Zhang, Dongmei},
  title = {Assess and Summarize: Improve Outage Understanding with Large Language Models},
  year = {2023},
  isbn = {9798400703270},
  publisher = {Association for Computing Machinery},
  address = {New York, NY, USA},
  url = {https://doi.org/10.1145/3611643.3613891},
  doi = {10.1145/3611643.3613891},
  booktitle = {Proceedings of the 31st ACM Joint European Software Engineering Conference and Symposium on the Foundations of Software Engineering},
  pages = {1657--1668},
  numpages = {12},
  location = {San Francisco, CA, USA},
  series = {ESEC/FSE 2023}
}

@inproceedings{jiajun2020tsg,
  author = {Jiang, Jiajun and Lu, Weihai and Chen, Junjie and Lin, Qingwei and Zhao, Pu and Kang, Yu and Zhang, Hongyu and Xiong, Yingfei and Gao, Feng and Xu, Zhangwei and Dang, Yingnong and Zhang, Dongmei},
  title = {How to mitigate the incident? an effective troubleshooting guide recommendation technique for online service systems},
  year = {2020},
  isbn = {9781450370431},
  publisher = {Association for Computing Machinery},
  address = {New York, NY, USA},
  url = {https://doi.org/10.1145/3368089.3417054},
  doi = {10.1145/3368089.3417054},
  booktitle = {Proceedings of the 28th ACM Joint Meeting on European Software Engineering Conference and Symposium on the Foundations of Software Engineering},
  pages = {1410--1420},
  numpages = {11},
  location = {Virtual Event, USA},
  series = {ESEC/FSE 2020}
}

@inproceedings{roy2024rca,
  author = {Roy, Devjeet and Zhang, Xuchao and Bhave, Rashi and Bansal, Chetan and Las-Casas, Pedro and Fonseca, Rodrigo and Rajmohan, Saravan},
  title = {Exploring {LLM}-Based Agents for Root Cause Analysis},
  year = {2024},
  isbn = {9798400706585},
  publisher = {Association for Computing Machinery},
  address = {New York, NY, USA},
  url = {https://doi.org/10.1145/3663529.3663841},
  doi = {10.1145/3663529.3663841},
  booktitle = {Companion Proceedings of the 32nd ACM International Conference on the Foundations of Software Engineering},
  pages = {208--219},
  numpages = {12},
  location = {Porto de Galinhas, Brazil},
  series = {FSE 2024}
}

@techreport{zhang2024flash,
  title={{FLASH}: A Workflow Automation Agent for Diagnosing Recurring Incidents},
  author={Zhang, Xuchao and Mittal, Tanish and Bansal, Chetan and Wang, Rujia and Ma, Minghua and Ren, Zhixin and Huang, Hao and Rajmohan, Saravan},
  year={2024},
  month={October},
  institution={Microsoft Research},
  url={https://www.microsoft.com/en-us/research/publication/flash-a-workflow-automation-agent-for-diagnosing-recurring-incidents/}
}

@inproceedings{zexin2024comet,
  author={Wang, Zexin and Li, Jianhui and Ma, Minghua and Li, Ze and Kang, Yu and Zhang, Chaoyun and Bansal, Chetan and Chintalapati, Murali and Rajmohan, Saravan and Lin, Qingwei and Zhang, Dongmei and Pei, Changhua and Xie, Gaogang},
  booktitle={2024 IEEE 35th International Symposium on Software Reliability Engineering (ISSRE)},
  title={Large Language Models Can Provide Accurate and Interpretable Incident Triage},
  year={2024},
  pages={523--534},
  doi={10.1109/ISSRE62328.2024.00056}
}

@misc{xie2024cloudatlas,
  title={Cloud Atlas: Efficient Fault Localization for Cloud Systems using Language Models and Causal Insight},
  author={Zhiqiang Xie and Yujia Zheng and Lizi Ottens and Kun Zhang and Christos Kozyrakis and Jonathan Mace},
  year={2024},
  eprint={2407.08694},
  archivePrefix={arXiv},
  primaryClass={cs.DC},
  doi={10.48550/arXiv.2407.08694},
  url={https://arxiv.org/abs/2407.08694}
}

@inproceedings{liu2025contextgaps,
  title={Bridging Context Gaps: Leveraging Coreference Resolution for Long Contextual Understanding},
  author={Liu, Yanming and Peng, Xinyue and Cao, Jiannan and Bo, Shi and Shen, Yanxin and Du, Tianyu and Cheng, Sheng and Wang, Xun and Yin, Jianwei and Zhang, Xuhong},
  booktitle={International Conference on Learning Representations (ICLR)},
  year={2025},
  url={https://openreview.net/forum?id=cPozlf9OaF}
}

@inproceedings{wang_executable_2024,
  title = {Executable Code Actions Elicit Better {LLM} Agents},
  booktitle = {Proceedings of the 41st International Conference on Machine Learning (ICML)},
  series = {Proceedings of Machine Learning Research},
  volume = {235},
  pages = {50208--50232},
  author = {Wang, Xingyao and Chen, Yangyi and Yuan, Lifan and Zhang, Yizhe and Li, Yunzhu and Peng, Hao and Ji, Heng},
  year = {2024}
}

@inproceedings{an_nissist_2024,
  title = {Nissist: An Incident Mitigation Copilot based on Troubleshooting Guides},
  booktitle = {ECAI 2024 - 27th European Conference on Artificial Intelligence},
  series = {Frontiers in Artificial Intelligence and Applications},
  volume = {392},
  publisher = {IOS Press},
  author = {An, Kaikai and Yang, Fangkai and Lu, Junting and Li, Liqun and Ren, Zhixing and Huang, Hao and Wang, Lu and Zhao, Pu and Kang, Yu and Ding, Hua and Lin, Qingwei and Rajmohan, Saravan and Zhang, Dongmei and Zhang, Qi},
  year = {2024},
  pages = {4471--4474},
  doi = {10.3233/FAIA241032}
}

@inproceedings{wu2024autogen,
  title={{AutoGen}: Enabling Next-Gen {LLM} Applications via Multi-Agent Conversation},
  author={Qingyun Wu and Gagan Bansal and Jieyu Zhang and Yiran Wu and Beibin Li and Erkang Zhu and Li Jiang and Xiaoyun Zhang and Shaokun Zhang and Jiale Liu and Ahmed Hassan Awadallah and Ryen W White and Doug Burger and Chi Wang},
  year={2024},
  booktitle={Proceedings of the Conference on Language Modeling (COLM)},
  url={https://openreview.net/forum?id=uAjxFFing2}
}

@misc{xu_core_2024,
  title = {{CoRE}: {LLM} as Interpreter for Natural Language Programming, Pseudo-Code Programming, and Flow Programming of {AI} Agents},
  url = {http://arxiv.org/abs/2405.06907},
  doi = {10.48550/arXiv.2405.06907},
  publisher = {arXiv},
  author = {Xu, Shuyuan and Li, Zelong and Mei, Kai and Zhang, Yongfeng},
  year = {2024},
  note = {arXiv:2405.06907}
}

@inproceedings{chen_icm_2020,
  author = {Chen, Zhuangbin and Kang, Yu and Li, Liqun and Zhang, Xu and Zhang, Hongyu and Xu, Hui and Zhou, Yangfan and Yang, Li and Sun, Jeffrey and Xu, Zhangwei and Dang, Yingnong and Gao, Feng and Zhao, Pu and Qiao, Bo and Lin, Qingwei and Zhang, Dongmei and Lyu, Michael R.},
  title = {Towards intelligent incident management: why we need it and how we make it},
  year = {2020},
  isbn = {9781450370431},
  publisher = {Association for Computing Machinery},
  address = {New York, NY, USA},
  url = {https://doi.org/10.1145/3368089.3417055},
  doi = {10.1145/3368089.3417055},
  booktitle = {Proceedings of the 28th ACM Joint Meeting on European Software Engineering Conference and Symposium on the Foundations of Software Engineering},
  series = {ESEC/FSE 2020}
}

@article{tang2024nl2kql,
  title={{NL2KQL}: From Natural Language to Kusto Query},
  author={Xinye Tang and Amir H. Abdi and Jeremias Eichelbaum and Mahan Das and Alex Klein and Nihal Irmak Pakis and William Blum and Daniel L Mace and Tanvi Raja and Namrata Padmanabhan and Ye Xing},
  journal={arXiv preprint arXiv:2404.02933},
  year={2024},
  doi={10.48550/arXiv.2404.02933}
}

@inproceedings{wang2021how,
  author = {Wang, Weijing and Chen, Junjie and Yang, Lin and Zhang, Hongyu and Zhao, Pu and Qiao, Bo and Kang, Yu and Lin, Qingwei and Rajmohan, Saravanakumar and Gao, Feng and Xu, Zhangwei and Dang, Yingnong and Zhang, Dongmei},
  title = {How Long Will it Take to Mitigate this Incident for Online Service Systems?},
  booktitle = {Proceedings of the 32nd IEEE International Symposium on Software Reliability Engineering (ISSRE)},
  year = {2021},
  pages = {36--46},
  doi = {10.1109/ISSRE52982.2021.00017}
}

@article{liu2025surveynl2sql,
  title={A Survey of Text-to-{SQL} in the Era of {LLMs}: Where Are We, and Where Are We Going?},
  author={Liu, Xinyu and Shen, Shuyu and Li, Boyan and Ma, Peixian and Jiang, Runzhi and Zhang, Yuxin and Fan, Ju and Li, Guoliang and Tang, Nan and Luo, Yuyu},
  journal={IEEE Transactions on Knowledge and Data Engineering},
  volume={37},
  number={10},
  pages={5735--5754},
  year={2025},
  doi={10.1109/TKDE.2025.3592032}
}

@inproceedings{hallucinationkql25,
  title = {Hallucination Detection in Structured Query Generation via {LLM} Self-Debating},
  author = {Li, Miaoran and Chen, Jiangning and Xu, Minghua and Wang, Xiaolong},
  booktitle = {Findings of the Association for Computational Linguistics: EMNLP 2025},
  year = {2025},
  doi = {10.18653/v1/2025.findings-emnlp.873}
}

@inproceedings{liu2025nl2sql,
  title={{NL2SQL-Bugs}: A Benchmark for Detecting Semantic Errors in {NL2SQL} Translation},
  author={Liu, Xinyu and Shen, Shuyu and Li, Boyan and Tang, Nan and Luo, Yuyu},
  booktitle={Proceedings of the 31st ACM SIGKDD Conference on Knowledge Discovery and Data Mining},
  pages={5662--5673},
  year={2025},
  doi={10.1145/3711896.3737427}
}

@inproceedings{schick2023toolformer,
  title={Toolformer: Language models can teach themselves to use tools},
  author={Schick, Timo and Dwivedi-Yu, Jane and Dess{\`\i}, Roberto and Raileanu, Roberta and Lomeli, Maria and Hambro, Eric and Zettlemoyer, Luke and Cancedda, Nicola and Scialom, Thomas},
  booktitle={Advances in Neural Information Processing Systems 36 (NeurIPS)},
  pages={68539--68551},
  year={2023}
}

@article{cardelli1996type,
  title={Type systems},
  author={Cardelli, Luca},
  journal={ACM Computing Surveys (CSUR)},
  volume={28},
  number={1},
  pages={263--264},
  year={1996},
  publisher={ACM New York, NY, USA},
  doi={10.1145/234313.234418}
}

@inproceedings{stroustrup1986overview,
  title={An overview of {C++}},
  author={Stroustrup, Bjarne},
  booktitle={Proceedings of the 1986 SIGPLAN workshop on Object-oriented programming},
  pages={7--18},
  year={1986}
}

@book{eckel2003thinking,
  title={Thinking in {JAVA}},
  author={Eckel, Bruce},
  year={2003},
  publisher={Prentice Hall Professional}
}

@inproceedings{matsakis2014rust,
  title={The {Rust} language},
  author={Matsakis, Nicholas D and Klock, Felix S},
  booktitle={Proceedings of the 2014 ACM SIGAda annual conference on High integrity language technology},
  pages={103--104},
  year={2014},
  doi={10.1145/2663171.2663188}
}

@book{pierce2002types,
  title={Types and programming languages},
  author={Pierce, Benjamin C},
  year={2002},
  publisher={MIT press}
}

@article{welsh2001seda,
  title={{SEDA}: An architecture for well-conditioned, scalable internet services},
  author={Welsh, Matt and Culler, David and Brewer, Eric},
  journal={ACM SIGOPS operating systems review},
  volume={35},
  number={5},
  pages={230--243},
  year={2001},
  publisher={ACM New York, NY, USA},
  doi={10.1145/502055.502057}
}

@inproceedings{isard2007dryad,
  title={Dryad: distributed data-parallel programs from sequential building blocks},
  author={Isard, Michael and Budiu, Mihai and Yu, Yuan and Birrell, Andrew and Fetterly, Dennis},
  booktitle={Proceedings of the 2nd ACM SIGOPS/EuroSys European conference on computer systems 2007},
  pages={59--72},
  year={2007},
  doi={10.1145/1272996.1273005}
}

@article{dean2008mapreduce,
  title={{MapReduce}: simplified data processing on large clusters},
  author={Dean, Jeffrey and Ghemawat, Sanjay},
  journal={Communications of the ACM},
  volume={51},
  number={1},
  pages={107--113},
  year={2008},
  publisher={ACM New York, NY, USA},
  doi={10.1145/1327452.1327492}
}

@article{las-casas_llexus_2024,
  title = {{LLexus}: An {AI} Agent System for Incident Management},
  author = {Las-Casas, Pedro and Kumbhare, Alok and Fonseca, Rodrigo and Agarwal, Sharad},
  journal = {ACM SIGOPS Operating Systems Review},
  volume = {58},
  number = {1},
  pages = {42--49},
  year = {2024},
  publisher = {ACM},
  address = {New York, NY, USA},
  doi = {10.1145/3689051.3689056},
  url = {https://dl.acm.org/doi/10.1145/3689051.3689056}
}

\end{document}